\documentclass[10pt,twocolumn,twoside]{IEEEtran}
\usepackage{times}
\usepackage{graphicx}
\usepackage{subfigure}
\usepackage{amsmath}
\usepackage{tabularx}
\usepackage{amssymb}
\usepackage{mathrsfs}
\usepackage{algorithm}
\usepackage{algorithmic}
\usepackage{multirow}
\usepackage{bm}
\usepackage{url}
\newcommand{\ve}{\mathbf{e}}

\newcommand{\vpi}{\bm{\pi}}

\newcommand{\semean}{\bm{\mu}}
\newcommand{\sevar}{\bm{\phi}}

\newcommand{\vZ}{\mathbf{Z}}
\newcommand{\vS}{\mathbf{S}}
\newcommand{\vT}{\mathcal{T}}

\begin{document}
\title{Learning Super-Resolution Jointly from External and Internal Examples}


\author{Zhangyang Wang\dag, Yingzhen Yang\dag, Zhaowen Wang\ddag, Shiyu Chang\dag, Jianchao Yang\ddag, Thomas S. Huang\dag
 \\
\dag Beckman Institute, University of Illinois at Urbana-Champaign, Urbana, IL 61801, USA\\
\ddag Adobe Systems Inc, San Jose, CA 95110, USA\\
{\tt\small \{zwang119, yyang58\}@illinois.edu, zhawang@adobe.com,  chang87@illinois.edu, jiayang@adobe.com, t-huang1@illinois.edu}
}

\IEEEcompsoctitleabstractindextext{%
\begin{abstract}
Single image super-resolution (SR) aims to estimate a high-resolution (HR) image from a low-resolution (LR) input. Image priors are commonly learned to regularize the otherwise seriously ill-posed SR problem, either using external LR-HR pairs or internal similar patterns. We propose joint SR to adaptively combine the advantages of both external and internal SR methods. We define two loss functions using sparse coding based external examples, and epitomic matching based on internal examples, as well as a corresponding adaptive weight to automatically balance their contributions according to their reconstruction errors. Extensive SR results demonstrate the effectiveness of the proposed method over the existing state-of-the-art methods, and is also verified by our subjective evaluation study.
\end{abstract}

}

\maketitle


\IEEEdisplaynotcompsoctitleabstractindextext
\IEEEpeerreviewmaketitle


\section{Introduction}

Super-resolution (SR) algorithms aim to constructing a high-resolution (HR) image from one or multiple low-resolution (LR) input frames \cite{Park2003}. This problem is essentially ill-posed because much information is lost in the HR to LR degradation process. Thus SR has to refer to strong image priors, that range from the simplest analytical smoothness assumptions, to more sophisticated statistical and structural priors learned from natural images \cite{Fattal2007}, \cite{Lin2004}, \cite{Yang2012}, \cite{Fattal2010}. The most popular single image SR methods rely on example-based learning techniques. Classical example-based methods learn the mapping between LR and HR image patches, from a large and representative external set of image pairs, and is thus denoted as \textit{external SR}. Meanwhile, images generally possess a great amount of self-similarities; such a self-similarity property motivates a series of \textit{internal SR} methods. With much progress being made, it is recognized that external and internal SR methods each suffer from their certain drawbacks. However, their complementary properties inspire us to propose the \textit{joint super-resolution} (joint SR),  that adaptively utilizes both external and internal examples for the SR task. The contributions of this paper are multi-fold:
\begin{itemize}

\item We propose joint SR exploiting both external and internal examples, by defining an adaptive combination of different loss functions.

\item We apply \textit{epitomic matching} \cite{JojicFK03} to enforcing self-similarity in SR. Compared the the local nearest neighbor (NN) matching adopted in \cite{Fattal2010}, epitomic matching features more robustness to outlier features, as well as the ability to perform efficient non-local searching.

\item We carry out a human subjective evaluation survey to evaluate SR result quality based on visual perception, among several state-of-the-art methods. 


\end{itemize}


\section{A Motivation Study of Joint SR}

\begin{figure*}[htbp]
\centering
\begin{minipage}{0.24\textwidth}
\centering \subfigure[\textit{Train}, the groundtruth of \newline carriage region] {
\includegraphics[width=\textwidth]{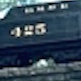}
}\end{minipage}
\begin{minipage}{0.24\textwidth}
\centering \subfigure[\textit{Train}, carriage region by \cite{Yang2012} \newline PSNR = 24.91 dB, SSIM = 0.7915] {
\includegraphics[width=\textwidth]{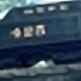}
}\end{minipage}
\begin{minipage}{0.24\textwidth}
\centering \subfigure [ \textit{Train}, carriage region by \cite{Fattal2010} \newline PSNR = 24.13 dB, SSIM = 0.8085] {
\includegraphics[width=\textwidth]{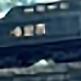}
}\end{minipage}
\\
\begin{minipage}{0.24\textwidth}
\centering \subfigure[\textit{Train}, the groundtruth of \newline brick region] {
\includegraphics[width=\textwidth]{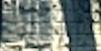}
}\end{minipage}
\begin{minipage}{0.24\textwidth}
\centering \subfigure [\textit{Train}, brick region by \cite{Yang2012} \newline PSNR = 18.84 dB, SSIM = 0.6576] {
\includegraphics[width=\textwidth]{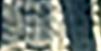}
}\end{minipage}
\begin{minipage}{0.24\textwidth}
\centering \subfigure [\textit{Train}, brick region by \cite{Fattal2010} \newline PSNR = 19.78 dB, SSIM = 0.7037] {
\includegraphics[width=\textwidth]{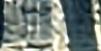}
}\end{minipage}
\\
\begin{minipage}{0.24\textwidth}
\centering \subfigure[\textit{Kid}, the groundtruth of  \newline left eye region] {
\includegraphics[width=\textwidth]{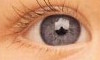}
}\end{minipage}
\begin{minipage}{0.24\textwidth}
\centering \subfigure[\textit{Kid}, left eye region by \cite{Yang2012} \newline PSNR = 22.43 dB, SSIM = 0.6286] {
\includegraphics[width=\textwidth]{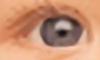}
}\end{minipage}
\begin{minipage}{0.24\textwidth}
\centering \subfigure [\textit{Kid}, left eye region by \cite{Fattal2010} \newline PSNR = 22.18 dB, SSIM = 0.5993] {
\includegraphics[width=\textwidth]{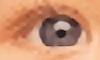}
}\end{minipage}
\\
\begin{minipage}{0.24\textwidth}
\centering \subfigure [\textit{Kid}, the groundtruth of \newline sweater region] {
\includegraphics[width=\textwidth]{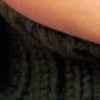}
}\end{minipage}
\begin{minipage}{0.24\textwidth}
\centering \subfigure [\textit{Kid}, sweater region by \cite{Yang2012} \newline PSNR = 24.16 dB, SSIM = 0.5444] {
\includegraphics[width=\textwidth]{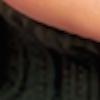}
}\end{minipage}
\begin{minipage}{0.24\textwidth}
\centering \subfigure [ \textit{Kid}, sweater region by \cite{Fattal2010} \newline PSNR = 24.45 dB, SSIM = 0.6018] {
\includegraphics[width=\textwidth]{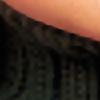}
}\end{minipage}
\caption{Visual comparisons of both external and internal SR methods on different image local regions. The PSNR and SSIM values are also calculated and reported.}
\label{fig:1}
\end{figure*}

\subsection{Related Work}
The joint utilization of both external and internal examples has been most studied for image denoising \cite{Zontak2011}. Mosseri et. al. \cite{Irani} first proposed that some image patches inherently prefer internal examples for denoising, whereas other patches inherently prefer external denoising. Such a preference is in essence the tradeoff between noise-fitting versus signal-fitting. Burger et. al. \cite{Harold} proposed a learning-based approach that automatically combines denoising results from an internal and an external method. The learned combining strategy outperforms both internal and external approaches across a wide range of images, being closer to theoretical bounds.

In SR literature, while the most popular methods are based on either external or internal similarities, there have been limited efforts to utilize one to regularize the other. The authors in \cite{Dong} incorporated both a local autoregressive (AR) model and a nonlocal self-similarity regularization term, into the sparse representation framework, weighted by constant coefficients. Yang et. al. \cite{Yang2013} learned the (approximated) nonlinear SR mapping function from a collection of external images with the help of in-place self-similarity. More recently, an explicitly joint model is put forward in \cite{WACV}, including two loss functions by sparse coding and local scale invariance, bound by an indicator function to decide which loss function will work for each patch of the input image. Despite the existing efforts, there is little understanding on how the external and internal examples interact with each other in SR, how to judge the external versus internal preference for each patch, and how to make them collaborate towards an overall optimized performance. 

External SR methods use a universal set of example patches to predict the missing (high-frequency) information for the HR image. In \cite{NN},  during the training phase, LR-HR patch pairs are collected. Then in the test phase, each input LR patch is found with a nearest neighbor (NN) match in the LR patch pool, and its corresponding HR patch is selected as the output. It is further formulated as a kernel ridge regression (KRR) in \cite{RG}. More recently, a popular class of external SR methods are associated with the \textit{sparse coding} technique \cite{Lee2007}, \cite{Yang2010}. The patches of a natural image can be represented as a sparse linear combination of elements within a redundant pre-trained dictionary. Following this principle, the advanced \textit{coupled sparse coding} is further proposed in  \cite{Yang2012}, \cite{Yang2010}. External SR methods are known for their capabilities to produce plausible image appearances. However, there is no guarantee that an arbitrary input patch can be well matched or represented by the external dataset of limited size. When dealing with some unique features that rarely appear in the given dataset, external SR methods are prone to produce either noise or oversmoothness \cite{Huang2010}. It constitutes the inherent problem of any external SR method with a finite-size training set \cite{Dong2012}. 

Internal SR methods search for example patches from the input image itself, based on the fact that patches often tend to recur within the image \cite{Glasner2009}, \cite{Mairal2009}, \cite{Huang2010}, or across different image scales \cite{Fattal2010}. Although internal examples provide a limited number of references, they are very relevant to the input image. However, this type of approach has a limited performance, especially for irregular patches without any discernible repeating pattern \cite{DD}. Due to the insufficient patch pairs, the mismatches of internal examples often lead to more visual artifacts. In addition, epitome was proposed in \cite{JojicFK03, Ni09, ChuYLCH10} to summarize both local and non-local similar patches and reduces the artifacts caused by neighborhood matching. We apply epitome as an internal SR technique in this paper, and evidence its advantages by our experiments.


\subsection{Comparing External and Internal SR Methods}

Both external and internal SR methods have different advantages and drawbacks. See Fig.~\ref{fig:1} for a few specific examples. The first two rows of images are cropped from the 3$\times$ SR results of the \textit{Train} image, and the last two rows from the 4$\times$ SR results of the \textit{Kid} image. Each row of images are cropped from the same spatial location of the groundtruth image, the SR result by the external method \cite{Yang2012}, and the SR result by the internal method \cite{Fattal2010}, respectively. In the first row, the top contour of carriage (c) contains noticeable structural deformations, and the numbers ``425'' are more blurred than those in (b). That is because the numbers can more easily find counterparts or similar structure components from an external dataset; but within the same image, there are few recurring patterns that look visually similar to the numbers. Internal examples generate sharper SR results in images (f) than (e), since the bricks repeat their own patterns frequently, and thus the local neighborhood is rich in internal examples. Another winning case of external examples is between (h) and (i), as in the latter, inconsistent artifacts along the eyelid and around the eyeball are obvious. Because the eye region is composed of complex curves and fine structures,  external examples encompass more suitable reference patches and perform a more natural-looking SR. In contrast, the repeating sweater textures lead to a sharper SR in (l) than that in (k). The PSNR and SSIM \cite{SSIM} results are also calculated for all, which further validate our visual observations.



These comparisons display the generally different, even complementary behaviors of external and internal SR. Based on the observations, we expect that the external examples contribute to visually pleasant SR results for smooth regions as well as some irregular structures that barely recur in the input. Meanwhile, internal examples serve as a powerful source to reproduce unique and singular features that rarely appear externally but repeat in the input image (or its different scales). Note that similar arguments have been validated statistically in the the image denoising literature \cite{Harold}.

\section{A Joint SR model}
Let $\mathbf{X}$ denote the HR image to be estimated from the LR input $\mathbf{Y}$. $\mathbf{X}_{ij}$ and $\mathbf{Y}_{ij}$ stand for the $(i,j)$-th ($i, j = 1, 2...$) patch from $\mathbf{X}$ and $\mathbf{Y}$, respectively.  Considering almost all SR methods work on patches, we define two loss functions $\ell_\mathcal{G} (\cdot)$ and $\ell_\mathcal{I} (\cdot)$ in a patch-wise manner, which enforce the external and internal similarities, respectively. While one intuitive idea is to minimize a weighted combination of the two loss functions, a patch-wise (adaptive) weight $\omega(\cdot) $ is needed to balance them.  We hereby write our proposed joint SR in the general form:
\begin{equation}
\begin{array}{l}\label{jsr}
\underset{\mathbf{X}_{ij}, \Theta_G, \Theta_I}{\min} \quad \ell_\mathcal{G} (\mathbf{X}_{ij}, \Theta_G|\mathbf{Y}_{ij}) + \omega(\Theta_G, \Theta_I|\mathbf{Y}_{ij}) \ell_\mathcal{I}(\mathbf{X}_{ij},  \Theta_I|\mathbf{Y}_{ij}).
\end{array}
\end{equation}
$\Theta_G$ and $\Theta_I$ are the latent representations of $\mathbf{X}_{ij}$ over the spaces of external and self examples, respectively. The form $f(\mathbf{X}_{ij}, \Theta|\mathbf{Y}_{ij})$, $f$ being $\ell_\mathcal{G}$, $\ell_\mathcal{I}$ or  $\omega$, represents the function dependent on variables $\mathbf{X}_{ij}$ and $\Theta$ ($\Theta_G$ or $\Theta_I$), with $\mathbf{Y}_{ij}$ known (we omit $\mathbf{Y}_{ij}$ in all formulations hereinafter). We will discuss each component in (\ref{jsr}) next. 




One specific form of joint SR will be discussed in this paper. However, note that with different choices of $\ell_\mathcal{G} \mathbf{(\cdot)}$, $\ell_\mathcal{I} \mathbf{(\cdot)}$, and $\omega(\cdot)$,  a variety of methods can be accommodated in the framework. For example, if we set $\ell_\mathcal{G} \mathbf{(\cdot)}$ as the (adaptively reweighted) sparse coding term, while choosing $\ell_\mathcal{I} \mathbf{(\cdot)}$ equivalent to the two local and non-local similarity based terms, then (\ref{jsr}) becomes the model proposed in \cite{Dong}, with $\omega(\cdot)$ being some empirically chosen constants. 

\subsection{Sparse Coding for External Examples}
The HR and LR patch spaces \{$\mathbf{X}_{ij}$\} and \{$\mathbf{Y}_{ij}$\} are assumed to be tied by some mapping function. With a well-trained coupled dictionary pair ($\mathbf{D_h}$, $\mathbf{D_l}$) (see \cite{Yang2012} for details on training a coupled dictionary pair), the \textit{coupled sparse coding} \cite{Yang2010} assumes that ($\mathbf{X}_{ij}$, $\mathbf{Y}_{ij}$) tends to admit a common sparse representation $\mathbf{a}_{ij}$. Since $\mathbf{X}$ is unknown, Yang et. al. \cite{Yang2010} suggest to first infer the sparse code $\mathbf{a}^L_{ij}$ of {$\mathbf{Y}_{ij}$} with respect to $\mathbf{D_l}$, and then use it as an approximation of $\mathbf{a}^H_{ij}$ (the sparse code of {$\mathbf{X}_{ij}$} with respect to $\mathbf{D_h}$), to recover $\mathbf{X}_{ij} \approx \mathbf{D_h} \mathbf{a}^L_{ij}$. We set $\Theta_G = \mathbf{a}_{ij}$ and constitute the loss function enforcing external similarity:
\begin{equation}
\begin{array}{l}\label{lg}
\ell_\mathcal{G} (\mathbf{X}_{ij}, \mathbf{a}_{ij}) = 
\lambda || \mathbf{a}_{ij}||_1 + || \mathbf{D_l}  \mathbf{a} _{ij}-  \mathbf{Y}_{ij}||_F^2 +  || \mathbf{D_h}  \mathbf{a}_{ij} -  \mathbf{X}_{ij}||_F^2.
\end{array}
\end{equation}


\subsection{Epitomic Matching for Internal Examples}

\subsubsection{The High Frequency Transfer Scheme}
Based on the observation that singular features like edges and corners in small patches tend to repeat almost identically across different image scales, Freedman and Fattal \cite{Fattal2010} applied the ``high frequency transfer'' method to searching the high-frequency component for a target HR patch, by NN patch matching across scales. Defining a linear interpolation operator $\mathcal{U}$ and a downsampling operator $\mathcal{D}$, for the input LR image $\mathbf{Y}$, we first obtain its initial upsampled image $\mathbf X^{'E} = \mathcal{U} (\mathbf{Y})$, and a smoothed input image $\mathbf{Y'} = \mathcal{D} (\mathcal{U} (\mathbf{Y}))$. Given the smoothed patch $\mathbf X_{ij}^{'E}$, the missing high-frequency band of each unknown patch $\mathbf X_{ij}^{E}$ is predicted by first solving a NN matching (\ref{freedman}):
\begin{equation}
\begin{array}{l}\label{freedman}
(m, n) = \arg\min_{(m, n) \in \mathcal{W}_{ij}} \| \mathbf Y'_{mn} - \mathbf X_{ij}^{'E}\|_F^2,
\end{array}
\end{equation}
where $\mathcal{W}_{ij}$ is defined as a small local searching window on image $\mathbf{Y'}$. We could also simply express it as $(m, n) = f_{NN} (\mathbf X_{ij}^{'E}, \mathbf{Y}')$. With the co-located patch $\mathbf{Y}_{mn}$ from $\mathbf{Y}$, the high-frequency band $\mathbf{Y}_{mn} - \mathbf{Y'}_{mn}$ is pasted onto $\mathbf X_{ij}^{'E}$, i.e., $\mathbf X_{ij}^{E} = \mathbf X_{ij}^{'E} + \mathbf{Y}_{mn} - \mathbf{Y'}_{mn}$.


\subsubsection{EPI: Epitomic Matching for Internal SR}

The matching of $\mathbf X_{ij}^{'E}$ over the smoothed input image $\mathbf{Y'}$ makes the core step of the high frequency transfer scheme. However, the performance of NN matching (\ref{freedman}) is degraded with the presence of noise and outliers. Moreover, the NN matching in \cite{Fattal2010} is restricted to a local window for efficiency, which potentially accounts for some rigid artifacts.

Instead, we propose \textit{epitomic matching} to replace NN matching in the above frequency transfer scheme. As a generative model, epitome \cite{ChuYLCH10, Yang14} summarizes a large set of raw image patches into a condensed representation in a way similar to Gaussian Mixture Models. We first learn an epitome $\ve_{\mathbf{Y'}}$ from $\mathbf{Y'}$, and then match each $\mathbf X_{ij}^{'E}$ over $\ve_{\mathbf{Y'}}$ rather than $\mathbf{Y'}$ directly. Assume $(m, n) = f_{ept} (\mathbf X_{ij}^{'E}, \ve_{\mathbf{Y'}})$, where $ f_{ept}$ denotes the procedure of epitomic matching by $\ve_{\mathbf{Y'}}$. It then follows the same way as in \cite{Fattal2010}: $\mathbf X_{ij}^{E} = \mathbf X_{ij}^{'E} + \mathbf{Y}_{mn} - \mathbf{Y'}_{mn}$: the only difference here is the replacement of $f_{NN}$ with $f_{ept}$. The high-frequency transfer scheme equipped with epitomic matching can thus be applied to SR by itself as well, named \textit{EPI} for short, which will be included in our experiments in Section 4 and compared to the method using NN matching in \cite{Fattal2010}. 

Since $\ve_{\mathbf{Y'}}$ summarizes the patches of the entire $\mathbf{Y'}$, the proposed epitomic matching benefits from non-local patch matching. In the absence of self-similar patches in the local neighborhood, epitomic matching weights refer to non-local matches, thereby effectively reducing the artifacts arising from local matching \cite{Fattal2010} in a restricted small neighborhood.  In addition, note that each epitome patch summarizes a batch of similar raw patches in $\mathbf{Y'}$. For any patch $\mathbf{Y'}_{ij}$ that contains certain noise or outliers in $\mathbf{Y'}$, its has a small posterior and thus tends not be selected as candidate matches for $\mathbf X_{ij}^{'E}$, improving the robustness of matching. The algorithm details of epitomic matching are included in Appendix.

Moreover, we can also incorporate Nearest Neighbor (NN) matching to our epitomic matching, leading to a enhanced patch matching scheme that features both non-local (by epitome) and local (by NN) matching. Suppose the high frequency components obtained by epitomic matching and NN matching for patch $\mathbf X_{ij}^{'E}$ are ${\mathbf H}_{ij,\ve}$ and ${\mathbf H}_{ij,{\rm NN}}$ respectively, we use a smart weighted average of the two as the final high frequency component ${\mathbf H}_{ij}$:
\begin{align}\label{eq:weighted-frequency}
&{\mathbf H}_{ij} = w{\mathbf H}_{ij,\ve}+(1-w){\mathbf H}_{ij,{\rm NN}}
\end{align}
where the weight $w=p( {\vT^*_{ij}|{\mathbf X_{ij}^{'E}},\ve })$ denotes the probability of the most probable hidden mapping given the patch $\mathbf X_{ij}^{'E}$. A higher $w$ indicates that the patch $\mathbf X_{ij}^{'E}$ is more likely to have a reliable match by epitomic matching (with the probability measured through the corresponding most probable hidden mapping), thereby a larger weight is associated with the epitomic matching, and vice versa. This is the practical implementation of EPI that we used in the paper.


Finally, we let $\Theta_I = \mathbf{X}_{ij}^E$ and define
\begin{equation}
\begin{array}{l}\label{li}
\ell_\mathcal{I} (\mathbf{X}_{ij}, \mathbf{X}_{ij}^E) = || \mathbf{X}_{ij} - \mathbf{X}_{ij}^E||_F^2,
\end{array}
\end{equation}
where $\mathbf{X}_{ij}^E$ is the internal SR result by epitomic matching.




\subsection{Learning the Adaptive Weights}
In \cite{Irani},  Mosseri et.al. showed that the internal versus external preference is tightly related to the Signal-to-Noise-Ratio (SNR) estimate of each patch. Inspired by that finding, we could seek similar definitions of "noise" in SR based on the latent representation errors. The \textit{external noise} is defined by the residual of sparse coding
\begin{equation}
\begin{array}{l}\label{ng}
N_g (\mathbf{a}_{ij}) =  || \mathbf{D_l}  \mathbf{a} _{ij}-  \mathbf{Y}_{ij}||_F^2.\\
\end{array}
\end{equation}
Meanwhile, the \textit{internal noise} finds its counterpart definition by the epitomic matching error within $f_{pet}$:
\begin{equation}
\begin{array}{l}\label{ni}
N_i (\mathbf{X}_{ij}^E) = || \mathbf Y'_{mn} - \mathbf X_{ij}^{'E}||_F^2,
\end{array}
\end{equation}
\noindent where $\mathbf Y'_{mn}$ is the matching patch in $\mathbf Y'$ for $\mathbf X_{ij}^{'E}$.

Usually, the two ``noises'' are on the same magnitude level, which aligns with the fact that external- and internal-examples will have similar performances on many (such as homogenous regions). However, there do exist patches where the two have significant differences in performances, as shown in Fig. \ref{fig:1}, which means the patch has a strong preference toward one of them. In such cases, the ``preferred'' term needs to be sufficiently emphasized. We thus construct the following patch-wise adaptive weight ($p$ is the hyperparameter):
\begin{equation}
\begin{array}{l}\label{C}
\omega (\mathbf{\alpha}_{ij}, \mathbf{X}_{ij}^E) = \exp(p\cdot [N_g (\mathbf{a}_{ij}) - N_i (\mathbf{X}^E_{ij}) ]).
\end{array}
\end{equation}
When the internal noise becomes larger, the weight decays quickly to ensure that external similarity dominates, and vice versa. 

\subsection{Optimization}
Directly solving (\ref{jsr}) is very complex due to the its high nonlinearity and entanglement among all variables. Instead, we follow the coordinate descent fashion \cite{Bertsekas1999} and solve the following three sub-problems iteratively.

\subsubsection{$\mathbf{a}_{ij}$-subproblem}
Fixing $\mathbf{X}_{ij}$ and  $\mathbf{X}_{ij}^E$, we have the following minimization w.r.t $ \mathbf{\alpha}_{ij}$
\begin{equation}
\begin{array}{l}\label{lg1}
\underset{\mathbf{a}_{ij}}{\min} \quad  \lambda ||\mathbf{a}_{ij}||_1 +   ||\mathbf{D_l} \mathbf{a}_{ij} - \mathbf{Y}_{ij}||_F^2 + ||\mathbf{D_h} \mathbf{a}_{ij} - \mathbf{X}_{ij}||_F^2  \\
\quad+ [\ell_\mathcal{I}(\mathbf{X}_{ij}, \mathbf{X}_{ij}^E) \cdot \exp( - p \cdot N_i (\mathbf{X}_{ij}^E)) ] \cdot \exp(p\cdot N_g (\mathbf{a}_{ij}) ).
\end{array}
\end{equation}
The major bottleneck of  exactly solving (\ref{lg1}) lies in the last exponential term. We let $\mathbf{a}^0_{ij}$ denote the $\mathbf{a}_{ij}$ value solved in the last iteration. We then apply first-order Taylor expansion to the last term of the objective in (\ref{lg1}), with regard to $N_g (\mathbf{\alpha}_{ij})$ at $\alpha_{ij}=\alpha^0_{ij}$, and solve the approximated problem as follows:
\begin{equation}
\begin{array}{l}\label{lg2}
\underset{\mathbf{a}_{ij}}{\min}  \quad \lambda ||\mathbf{a}_{ij}||_1 +  (1+C) ||\mathbf{D_l} \mathbf{a}_{ij} - \mathbf{Y}_{ij}||_F^2 + ||\mathbf{D_h} \mathbf{a}_{ij} - \mathbf{X}_{ij}||_F^2,
\end{array}
\end{equation}
where $C$ is the constant coefficient:
\begin{equation}
\begin{array}{l}\label{C}
C = [\ell_\mathcal{I}(\mathbf{X}_{ij}, \mathbf{X}_{ij}^E) \cdot \exp( - p \cdot N_i (\mathbf{X}_{ij}^E) ] \cdot [p\cdot \exp(p\cdot N_g (\mathbf{a}_{ij}^0)]\\
\quad  = p \ell_\mathcal{I}(\mathbf{X}_{ij}, \mathbf{X}_{ij}^E) \cdot \omega (\mathbf{\alpha^0}_{ij}, \mathbf{X}_{ij}^E).
\end{array}
\end{equation}
(\ref{lg2}) can be conveniently solved by the feature sign algorithm \cite{Lee2007}. Note (\ref{lg2}) is a valid approximation of (\ref{lg1}) since $\mathbf{a}_{ij}$ and $\mathbf{a}^0_{ij}$ become quite close after a few iterations, so that the higher-order Taylor expansions can be reasonably ignored. 

Another noticeable fact is that since $C>0$, the second term is always emphasized more than the third term, which makes sense as $\mathbf{Y}_{ij}$ is the ``accurate'' LR image, while $\mathbf{X}_{ij}$ is just an estimate of the HR image and is thus less weighted. Further considering the formulation (\ref{C}), $C$ grows up as $\omega (\mathbf{\alpha^0}_{ij}, \mathbf{X}_{ij}^E)$ turns larger. That implies when external SR becomes the major source of ``SR noise'' on a patch in the last iteration,  (\ref{lg2}) will accordingly rely less on the last solved $\mathbf{X}_{ij}$.

\subsubsection{$\mathbf{X}_{ij}^E$-subproblem}
Fixing $\mathbf{a}_{ij}$ and  $\mathbf{X}_{ij}$, the $\mathbf{X}_{ij}^E$-subproblem becomes
\begin{equation}
\begin{array}{l}\label{xe}
\underset{\mathbf{X}^E_{ij}}{\min} \quad \exp( -p \cdot || \mathbf Y'_{mn} - \mathbf X_{ij}^{'E}||_F^2) \ell_\mathcal{I}(\mathbf{X}_{ij},  \mathbf{X}^E_{ij}),
\end{array}
\end{equation}

While in Section III.B.2, $X_{i,j}^E$ is directly computed from the input LR image, the objective in (\ref{xe}) is dependent on not only $X_{i,j}^E$ but also $X_{i,j}$, which is not necessarily minimized by the best match $X_{i,j}^E$ obtained from solving $f_{ept}$. In our implementation, the $K$ best candidates ($K$ = 5) that yield minimum matching errors of solving $f_{ept}$ are first obtained.  Among all those candidates, we further select the one that minimizes the loss value as defined in (\ref{xe}). By this discrete search-type algorithm, $X_{i,j}^E$ becomes a latent variable to be updated together with $X_{i,j}$ per iteration, and is better suited for the global optimization than the simplistic solution by solving $f_{ept}$. 


\subsubsection{$\mathbf{X}_{ij}$-subproblem}
With both $\mathbf{a}_{ij}$ and  $\mathbf{X}_{ij}^E$ fixed, the solution of $\mathbf{X}_{ij}$ simply follows a weight least square (WLS) problem:
\begin{equation}
\begin{array}{l}\label{wls}
\underset{\mathbf{X}_{ij}}{\min} \quad   ||\mathbf{D_h} \mathbf{a}_{ij} - \mathbf{X}_{ij}||_F^2  + \omega(\mathbf{a}_{ij}, \mathbf{X}_{ij}^E) || \mathbf{X} - \mathbf{X}_{ij}^E||_F^2,
\end{array}
\end{equation}
with an explicit solution:
\begin{equation}
\begin{array}{l}\label{wls2}
\mathbf{X}_{ij} = \frac{\mathbf{D_h} \mathbf{a}_{ij} + \omega(\mathbf{\alpha}_{ij}, \mathbf{X}_{ij}^E)\cdot  \mathbf{X}_{ij}^E}{1+ \omega(\mathbf{a}_{ij}, \mathbf{X}_{ij}^E)}.
\end{array}
\end{equation}


\begin{figure*}[htbp]
\centering
\begin{minipage}{0.40\textwidth}
\centering \subfigure[BCI (PSNR = 25.29 dB, SSIM = 0.8762)] {
\includegraphics[width=\textwidth]{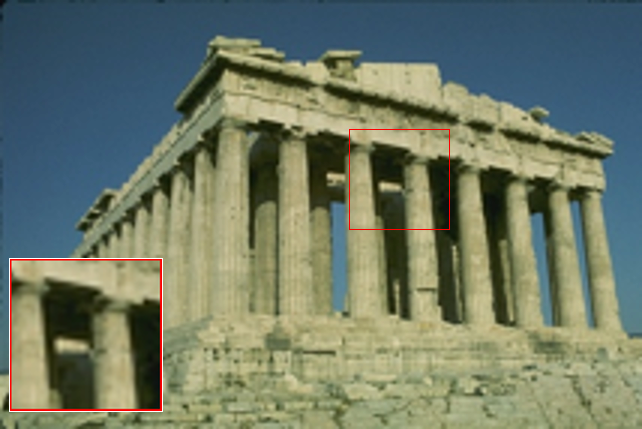}
}\\
\subfigure[CSC (PSNR = 26.20 dB, SSIM = 0.8924)] {
\includegraphics[width=\textwidth]{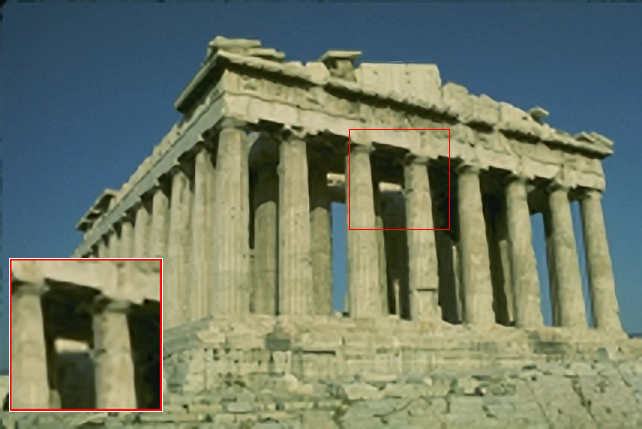}
}\\
\subfigure[LSE (PSNR = 21.17 dB, SSIM = 0.7954)] {
\includegraphics[width=\textwidth]{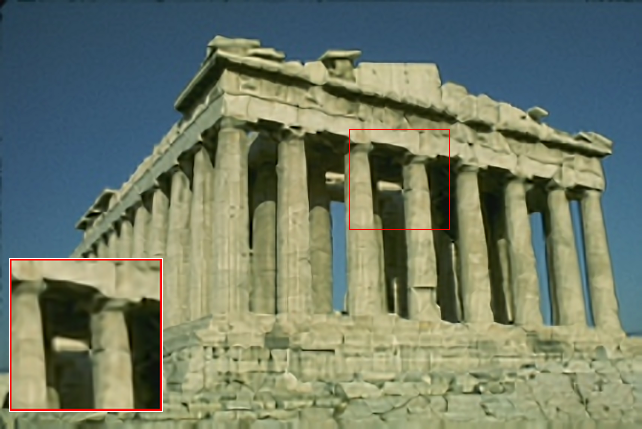}
}\\
\subfigure[EPI (PSNR = 24.34 dB, SSIM = 0.8901)] {
\includegraphics[width=\textwidth]{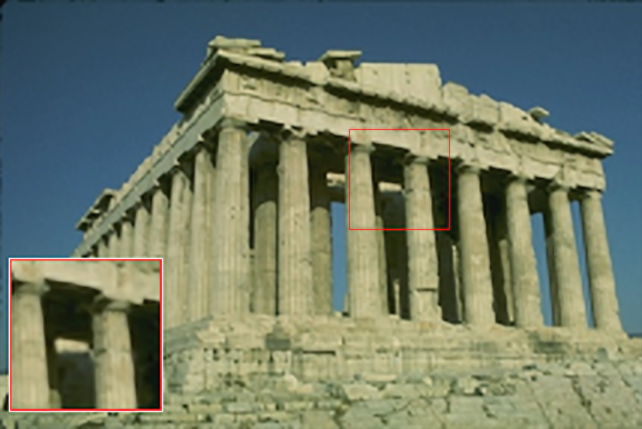}
}\\
\end{minipage}
\begin{minipage}{0.40\textwidth}
\centering \subfigure[IER (PSNR = 25.54 dB, SSIM = 0.8937)] {
\includegraphics[width=\textwidth]{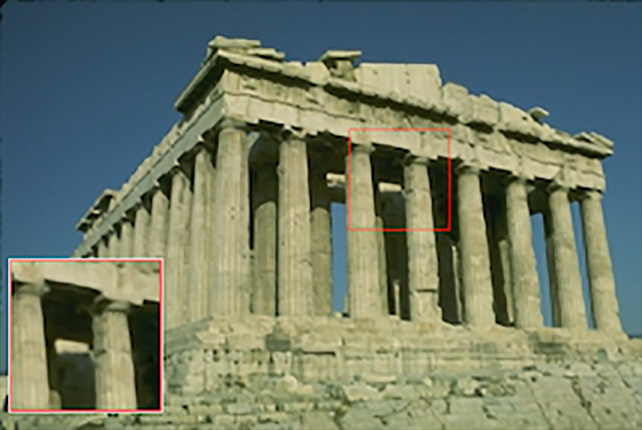}
}\\
\subfigure[JSR (PSNR = 27.87 dB, SSIM = 0.9327)] {
\includegraphics[width=\textwidth]{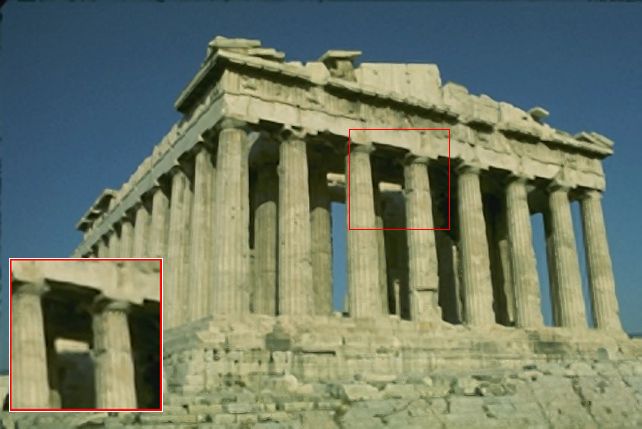}
}\\
\subfigure[Groundtruth] {
\includegraphics[width=\textwidth]{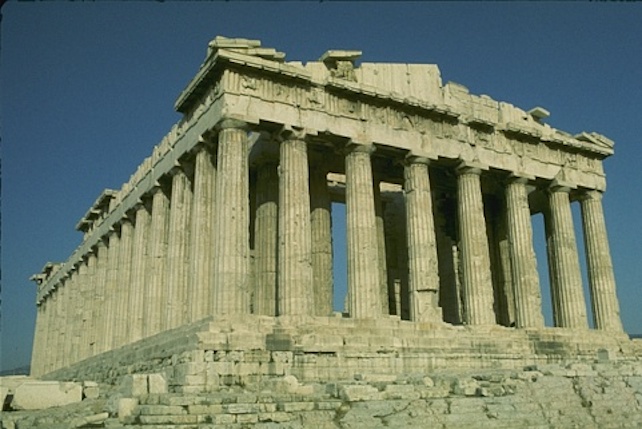}
}\\
\subfigure[LR input] {
\includegraphics[width=\textwidth]{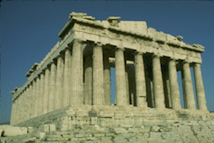}
}\end{minipage}
\caption{3$\times$ SR results of the \textit{Temple} image. }
\label{fig:temple}
\end{figure*}

\section{Experiments}

\subsection{Implementation Details}

We itemize the parameter and implementation settings for the following group of experiments:
\begin{itemize}
\item  We use $5 \times 5$ patches with one pixel overlapping for all experiments except those on SHD images in Section 4.4, where the patch size is  $25 \times 25$ with five pixel overlapping.
\item In (\ref{lg}), we adopt the $\mathbf{D_l}$ and $\mathbf{D_h}$ trained in the same way as in \cite{Yang2012}, due to the similar roles played by the dictionaries in their formulation and our $\ell_G$ function. However, we are aware that such $D_l$ and $D_h$ are not optimized for the proposed method, and will integrate a specifically designed dictionary learning part in future work. $\lambda$ is empirically set as 1.
\item In (\ref{li}), the size of the epitome is $\frac{1}{4}$ of the image size. 
\item In (\ref{C}), we set $p=1$ for all experiments. We also observed in experiments that a larger $p$ will usually lead to a faster decrease in objective value, but the SR result quality may degrade a bit.
\item  We initialize $\mathbf{a}_{ij}$ by solving coupled sparse coding in \cite{Yang2012}. $\mathbf{X}_{ij}$ is initialized by bicubic interpolation.
 \item We set the maximum iteration number to be 10 for the coordinate descent algorithm. For SHD cases, the maximum iteration number is adjusted to be 5.
 \item For color images, we apply SR algorithms to the illuminance channel only, as humans are more sensitive to illuminance changes. We then interpolate the color layers (Cb, Cr) using plain bi-cubic interpolation.
 \end{itemize}

\subsection{Comparison with State-of-the-Art Results}

We compare the proposed method with the following selection of competitive methods as follows,
\begin{itemize}
\item  \textit{Bi-Cubic Interpolation (``BCI'' for short and similarly hereinafter)}, as a comparison baseline.
\item \textit{Coupled Sparse Coding (CSC)} \cite{Yang2012}, as the classical external-example-based SR.
\item \textit{Local Self-Example based SR (LSE)} \cite{Fattal2010}, as the classical internal-example-based SR.
\item \textit{Epitome-based SR (EPI)}. We compare EPI to LSE to demonstrate the advantage of epitomic matching over the local NN matching.
\item \textit{SR based on In-place Example Regression (IER)} \cite{Yang2013}, as the previous SR utilizing both external and internal information.
\item \textit{The proposed joint SR (JSR)}.
\end{itemize}
We list the SR results (best viewed on a high-resolution display) for two test images: \textit{Temple} and \textit{Train}, by an amplifying factor of 3. PSNR and SSIM measurements, as well as zoomed local regions (using nearing neighbor interpolation), are available for different methods as well. 

\begin{figure*}[htbp]
\centering
\begin{minipage}{0.40\textwidth}
\centering \subfigure[BCI (PSNR = 26.14 dB, SSIM = 0.9403)] {
\includegraphics[width=\textwidth]{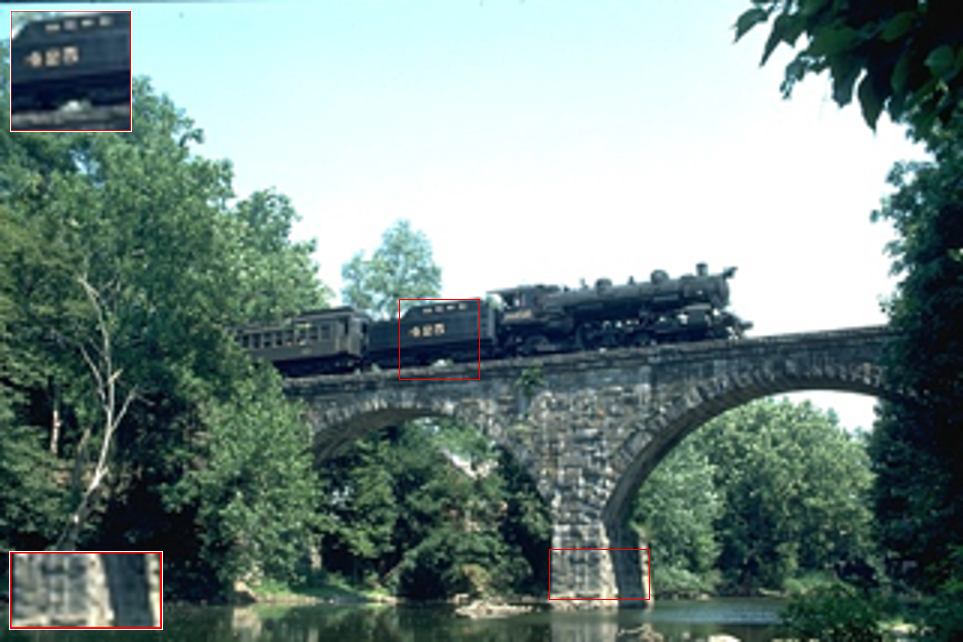}
}\\
\subfigure[CSC (PSNR = 26.58 dB, SSIM = 0.9506)] {
\includegraphics[width=\textwidth]{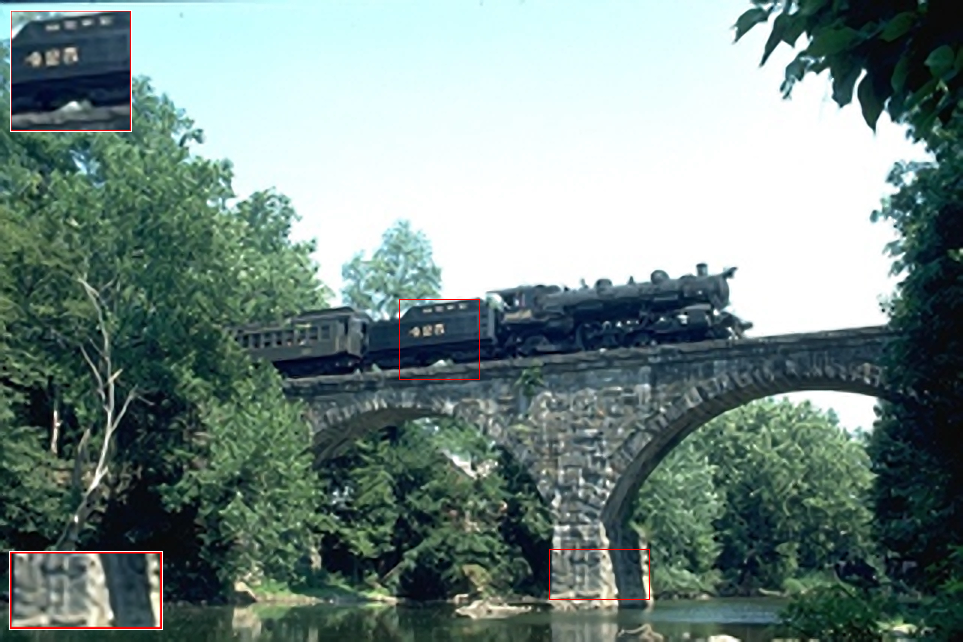}
}\\
\subfigure[LSE (PSNR = 22.54 dB, SSIM = 0.8850)] {
\includegraphics[width=\textwidth]{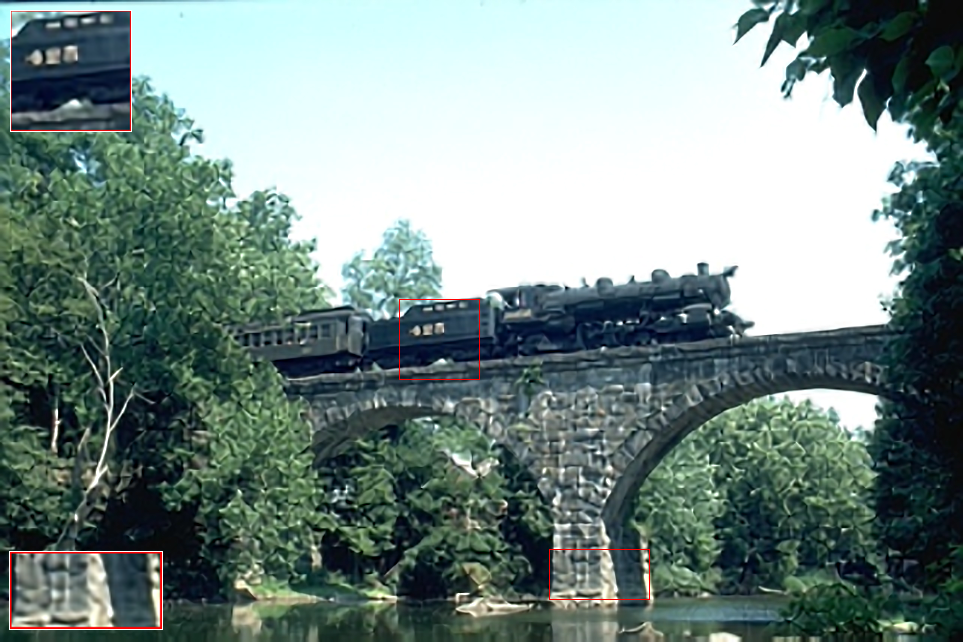}
}\\
\subfigure[EPI (PSNR = 26.22 dB, SSIM = 0.9487)] {
\includegraphics[width=\textwidth]{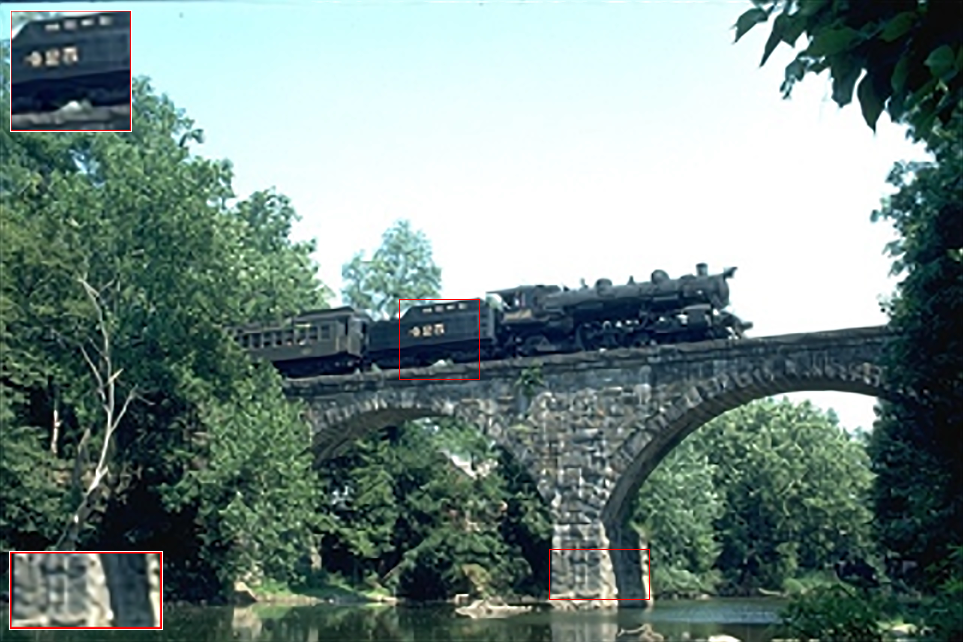}
}\end{minipage}
\begin{minipage}{0.40\textwidth}
\centering\subfigure[IER (PSNR = 24.80 dB, SSIM = 0.9323)] {
\includegraphics[width=\textwidth]{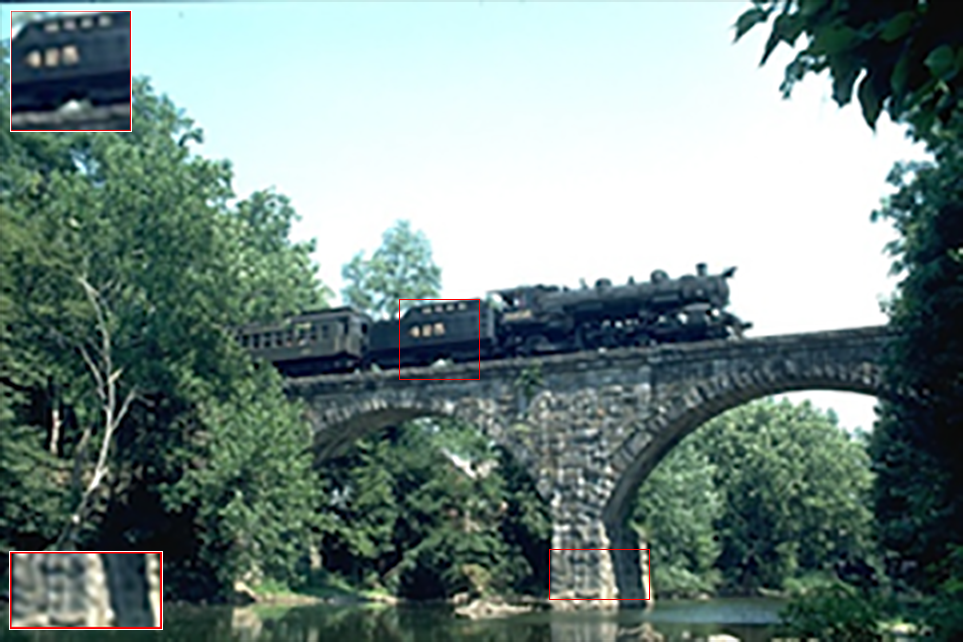}
}\\
\subfigure[JSR (PSNR = 28.02 dB, SSIM = 0.9796)] {
\includegraphics[width=\textwidth]{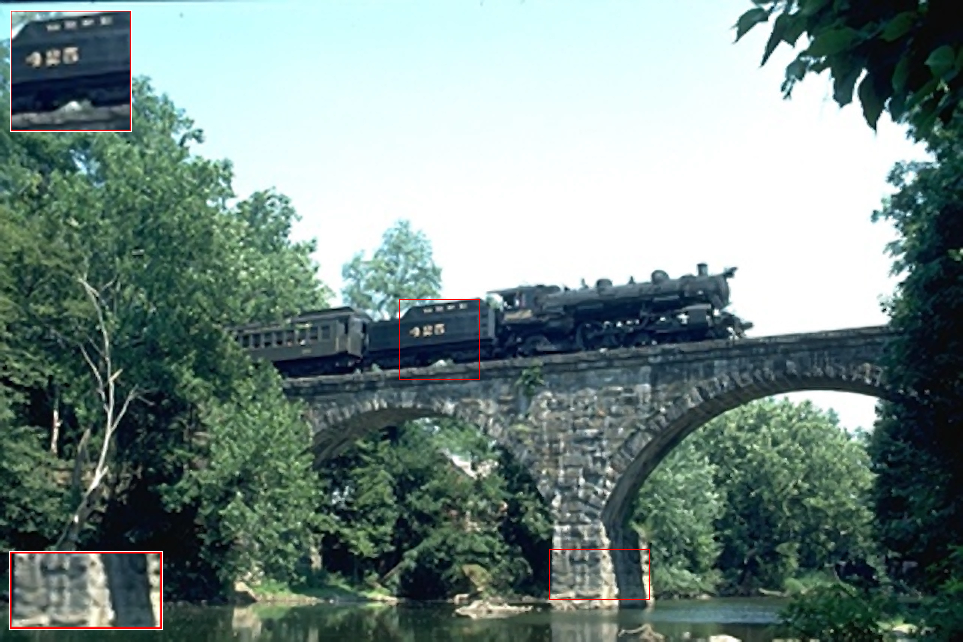}
}\\
\subfigure[Groundtruth] {
\includegraphics[width=\textwidth]{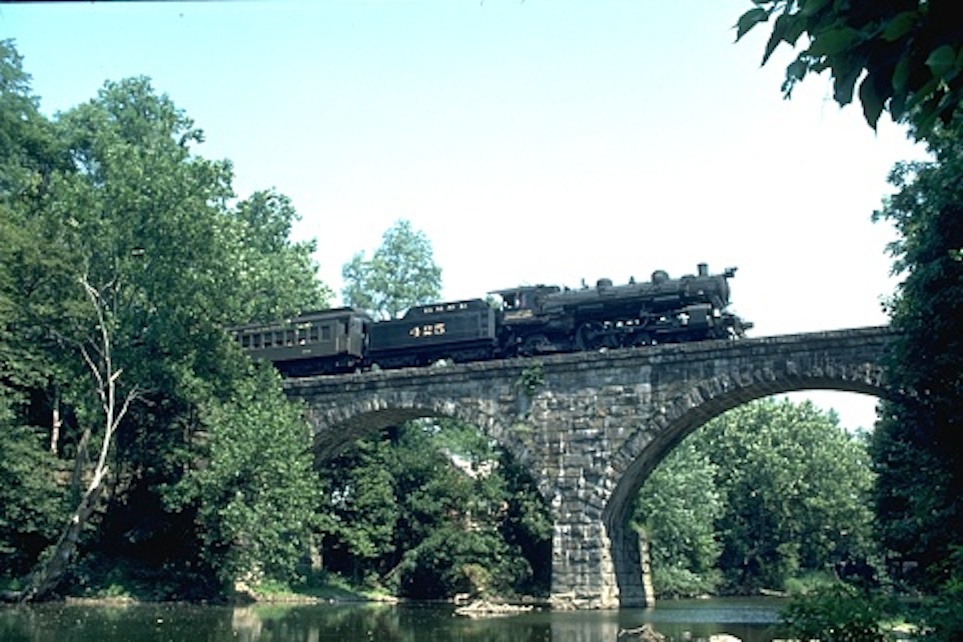}
}\\
\subfigure[LR input] {
\includegraphics[width=\textwidth]{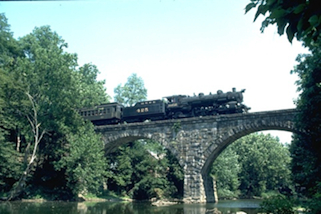}
}\end{minipage}
\caption{3$\times$ SR results of the \textit{Train} image. }
\label{fig:train}
\end{figure*}



In Fig. \ref{fig:temple}, although greatly outperforming the naive BCI, the external-example based CSC tends to lose many fine details. In contrast, LSE brings out an overly sharp SR result with observable blockiness. EPI produces a more visually pleasing result, through searching for the matches over the entire input efficiently by the pre-trained epitome rather than a local neighborhood. Therefore, EPI substantially reduces the artifacts compared to LSE. But without any external information available, it is still incapable of inferring enough high-frequency details from the input solely, especially under a large amplifying factor. The result of IER greatly improves but is still accompanied with occasional small artifacts. Finally, JSR provides a clear recovery of the steps, and it reconstructs the most pillar textures. In Fig. \ref{fig:train},  JSR is the only algorithm which clearly recovers the number on the carrier and the bricks on the bridge simultaneously. The performance superiorities of JSR are also verified by the PSNR comparisons, where larger margins are obtained by JSR over others in both cases.


Next, we move on to the more challenging 4$\times$ SR case, using the \textit{Chip} image which is quite abundant in edges and textures. Since we have no ground truth for the \textit{Chip} image of 4$\times$ size, only visual comparisons are presented. Given such a large SR factor, the CSC result is a bit blurry around the characters on the surface of chip. Both LSE and EPI create jaggy artifacts along the long edge of the chip, as well as small structure distortions. The IER result cause less artifacts but in sacrifice of detail sharpness. The JSR result presents the best SR with few artifacts. 

The key idea of JSR is utilizing the complementary behavior of both external and internal SR methods. Note when one inverse problem is better solved, it also makes a better parameter estimate for solving the other. JSR is not a simple static weighted average of external SR (CSC) and internal SR (EPI). When optimized jointly, the external and internal subproblems can "boost" each other (through auxiliary variables), and each performs better than being applied independently. That is why JSR gets details that exist in neither internal or external SR result.

To further verify the superiority of JSR numerically, we compare the average PSNR and SSIM results of a few recently-proposed, state-of-the-art single image SR methods, including CSC, LSE, the Adjusted Anchored Neighborhood Regression (A+) \cite{A+}, and the latest Super-Resolution Convolutional Neural Network (SRCNN) \cite{Tang}. Table \ref{set} reports the results on the widely-adopted Set 5 and Set 14 datasets, in terms of both PSNR and SSIM. First, it is not a surprise to us, that JSR does not always yield higher PSNR than SRCNN, et. al., as the epitomic matching component is not meant to be optimized under Mean-Square-Error (MSE) measure, in contrast to the end-to-end MSE-driven regression adopted in SRCNN. However, it is notable that JSR is particularly more favorable by SSIM than other methods, owing to the self-similar examples that convey input-specific structural details. Considering SSIM measures image quality more consistently with human perception, the observation is in accordance with our human subject evaluation results (see Section IV. E).

\begin{table*}[t]
\begin{center}
\caption{Average PSNR (dB) and SSIM performances comparisons on the Set 5 and Set 14 datasets}
\label{set}
\vspace{0.5em}
\begin{tabular}{|c|c|c|c|c|c|c|c|}
\hline
 & & Bicubic & Sparse Coding \cite{Yang2010} & Freedman et.al. \cite{Fattal2010}& A+ \cite{A+} & SRCNN \cite{Tang} &  JSR \\
\hline
$\multirow{2}{*}{\textit{Set 5, $s_t$=2}}$ &PSNR  & 33.66 & 35.27 & 33.61 & 36.24 & 36.66 & \textbf{36.71}\\
\cline{2-8}
$$ &SSIM & 0.9299 & 0.9540 & 0.9375 & 0.9544 & 0.9542 & \textbf{0.9573} \\
\hline
$\multirow{2}{*}{\textit{Set 5, $s_t$=3}}$ &PSNR & 30.39 & 31.42 & 30.77 & 32.59 & \textbf{32.75} & 32.54\\
\cline{2-8}
$$ &SSIM  & 0.8682 & 0.8821 & 0.8774 &0.9088 & 0.9090 & \textbf{0.9186} \\
\hline
$\multirow{2}{*}{\textit{Set 14, $s_t$=2}}$ &PSNR & 30.23 & 31.34 & 31.99 & \textbf{32.58} & 32.45& 32.54  \\
\cline{2-8}
$$ &SSIM & 0.8687 & 0.8928 & 0.8921 & 0.9056 & 0.9067 & \textbf{0.9082}  \\
\hline
$\multirow{2}{*}{\textit{Set 14, $s_t$=3}}$ &PSNR & 27.54 & 28.31 & 28.26 & 29.13 &\textbf{29.60} & 29.49  \\
\cline{2-8}
$$ &SSIM  & 0.7736 & 0.7954 & 0.8043 & 0.8188 & 0.8215 & \textbf{0.8242} \\
\hline
\end{tabular}
\end{center}
\end{table*}

\begin{figure*}[htbp]
\centering
\begin{minipage}{0.40\textwidth}
\centering \subfigure[BCI] {
\includegraphics[width=\textwidth]{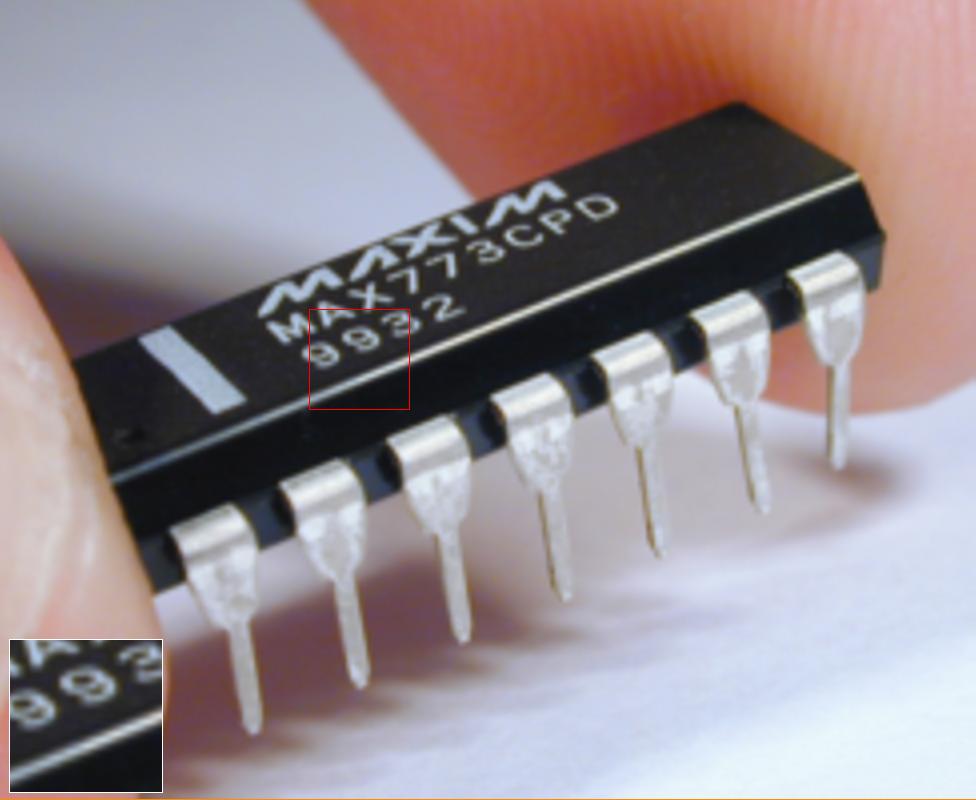}
}\\
\subfigure[CSC] {
\includegraphics[width=\textwidth]{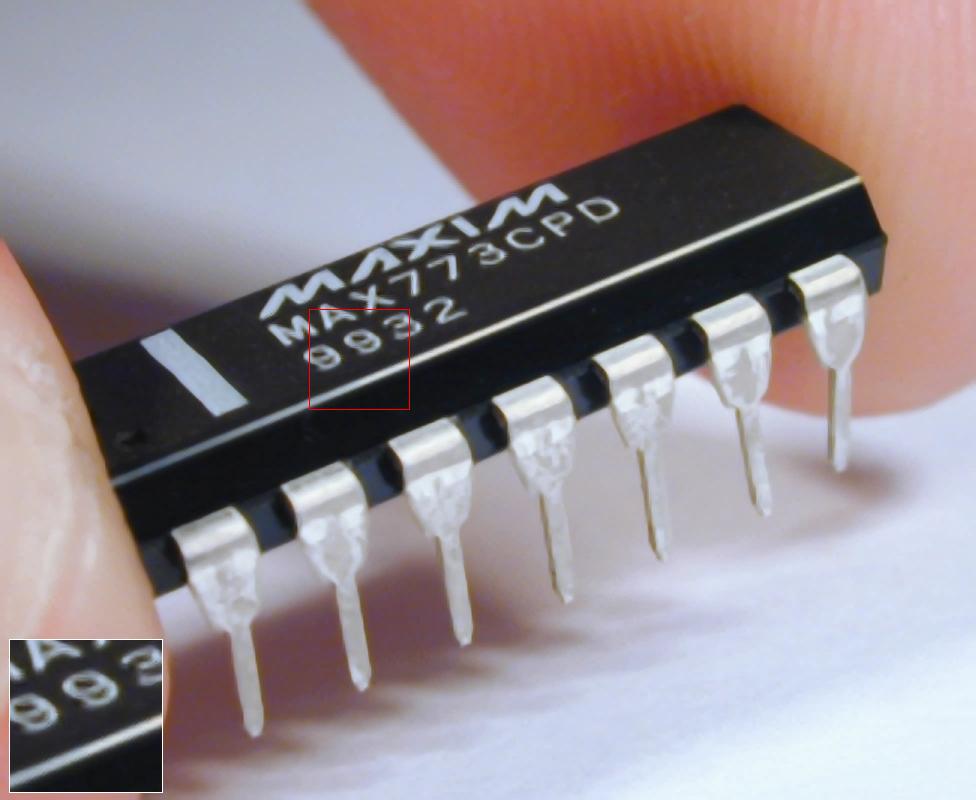}
}\\
\subfigure[LSE] {
\includegraphics[width=\textwidth]{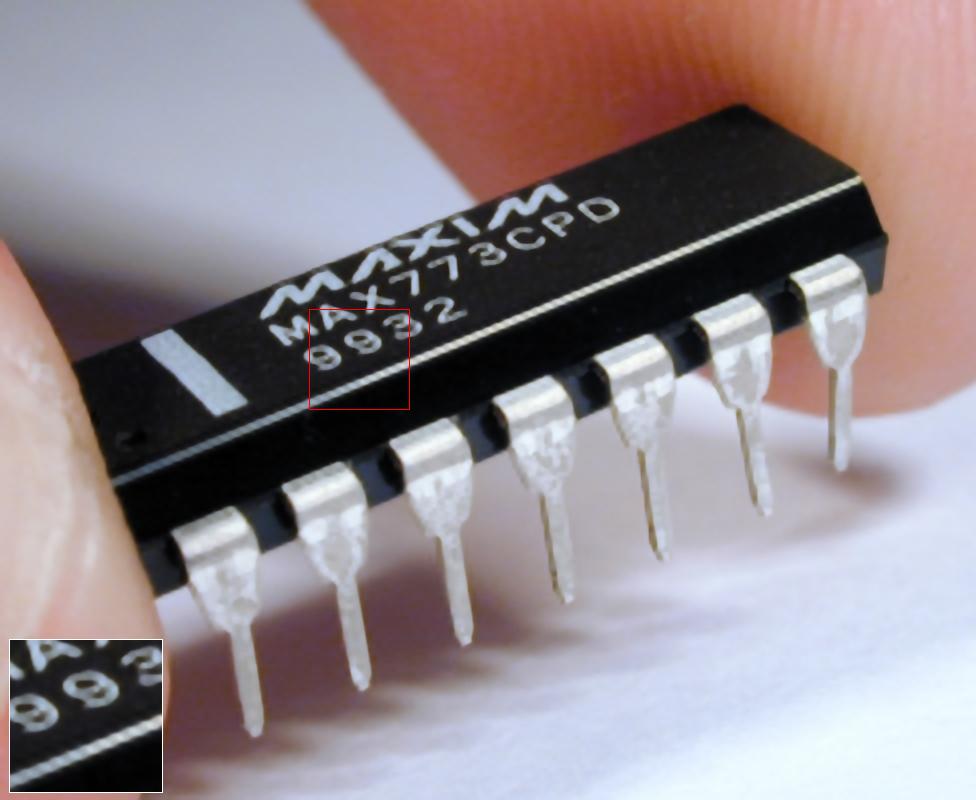}
}\end{minipage}
\begin{minipage}{0.40\textwidth}
\centering \subfigure[EPI] {
\includegraphics[width=\textwidth]{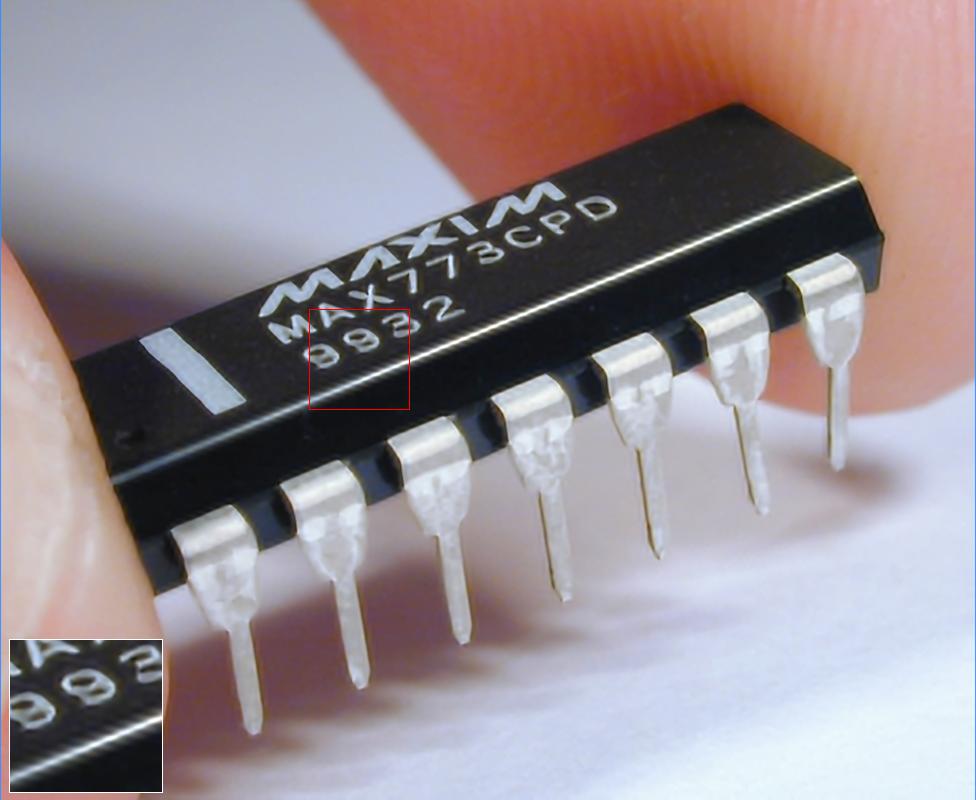}
}\\
\subfigure[IER] {
\includegraphics[width=\textwidth]{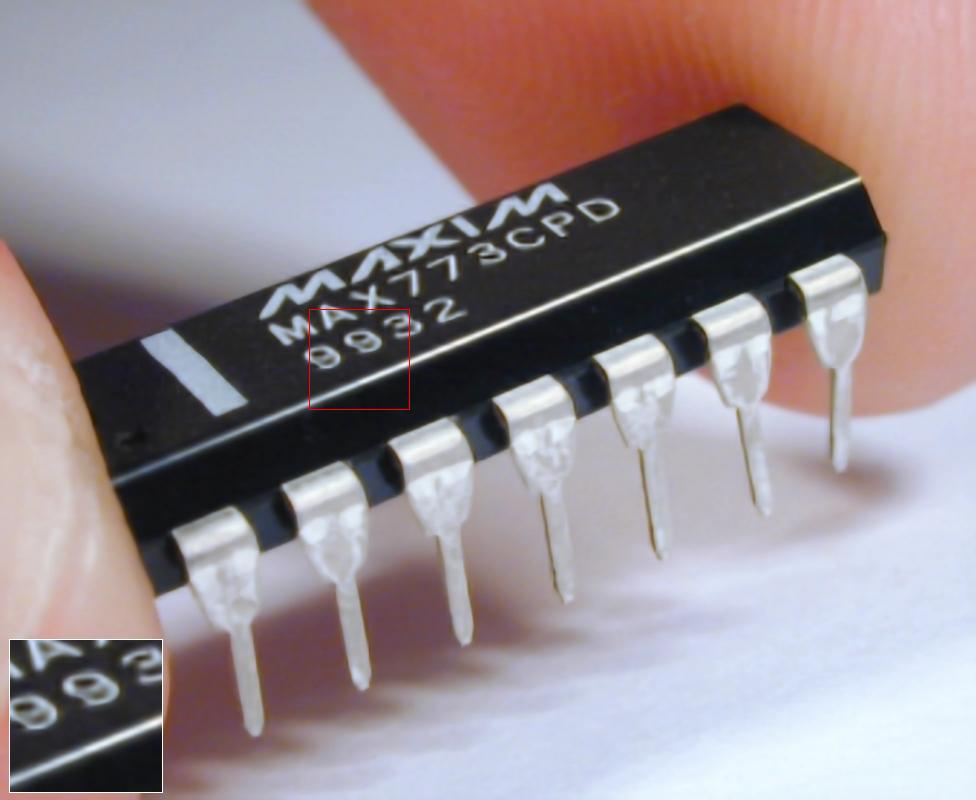}
}\\
\subfigure[JSR] {
\includegraphics[width=\textwidth]{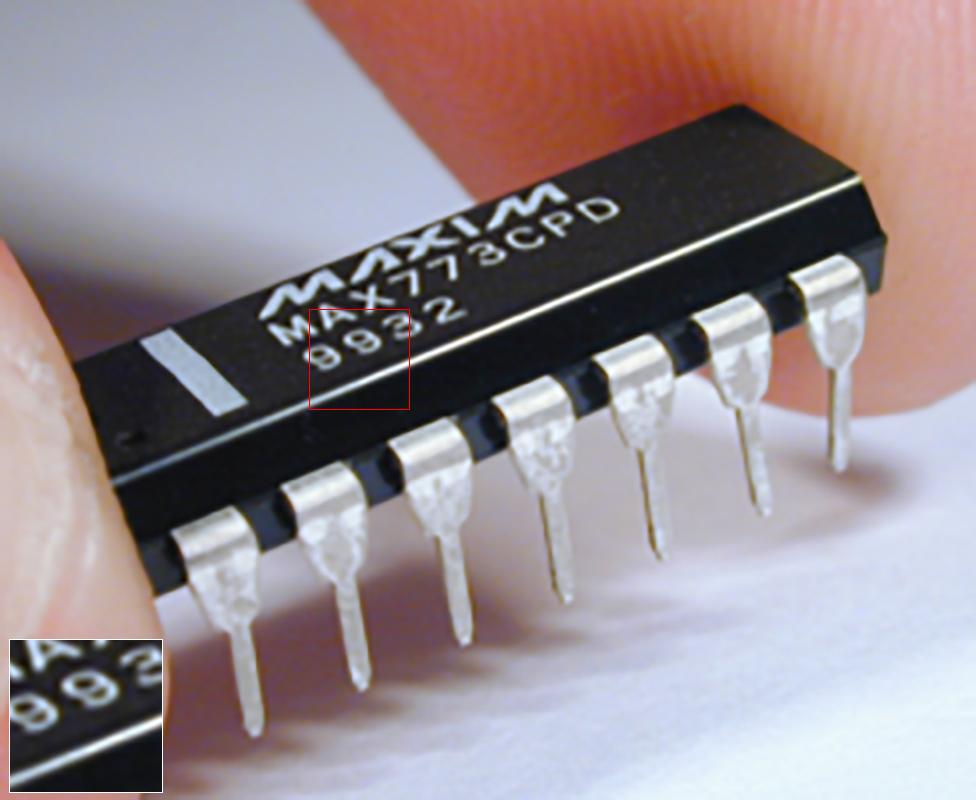}
}\end{minipage}
\centering \subfigure[LR input] {
\includegraphics[width=0.10\textwidth]{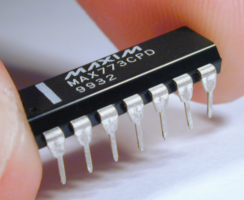}
}
\caption{4$\times$ SR results of the \textit{Chip} image.}
\label{fig:chip}
\end{figure*}

\begin{figure*}[htbp]
\centering
\begin{minipage}{0.32\textwidth}
\centering \subfigure[Temple] {
\includegraphics[width=\textwidth]{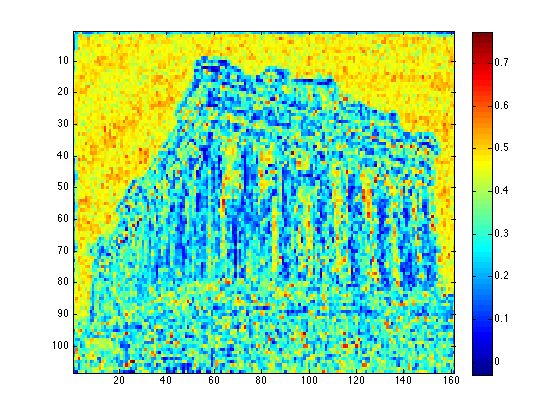}
}\end{minipage}
\begin{minipage}{0.32\textwidth}
\centering\subfigure[Train] {
\includegraphics[width=\textwidth]{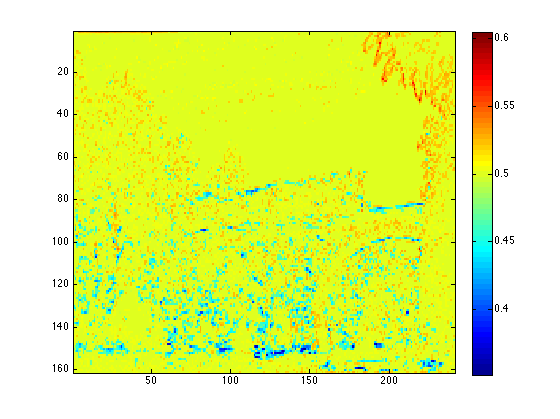}
}\end{minipage}
\begin{minipage}{0.32\textwidth}
\centering \subfigure[Chip] {
\includegraphics[width=\textwidth]{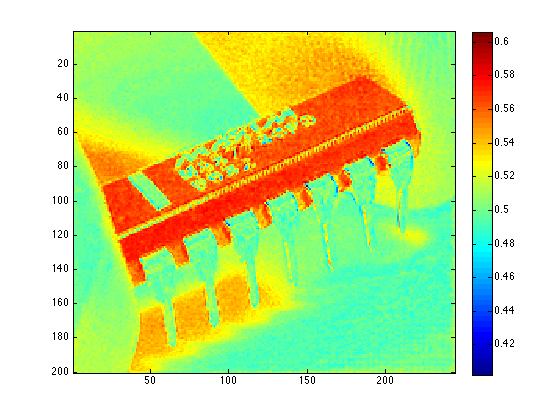}
}\end{minipage}
\caption{The weight maps of (a) \textit{Temple} image; (b) \textit{Train} image; (c) \textit{Chip} image.}
\label{fig:map}
\end{figure*}


%
%

 \subsection{Effect of Adaptive Weight}

To demonstrate how the proposed joint SR will benefit from the learned adaptive weight (\ref{C}), we compare 4$\times$ SR results of \textit{Kid} image, between joint SR solving (\ref{jsr}), and its counterpart with fixed global weights , i.e. set the weight $\omega$ as constant for all patches. Table 1 shows that the joint SR with an adaptive weight gains a consistent PSNR advantage over the SR with a large range of fixed weights. 

\begin{table}[t]
\begin{center}
\caption{The PSNR values (dB) with various fixed global weights (PSNR = 24.1734 dB with an adaptive weight)}
\vspace{0.1em}
\begin{tabular}{|c|c|c|c|c|}
\hline
$\omega$ = 0.1 & $\omega$ = 1 & $\omega$ = 3 & $\omega$ = 5 & $\omega$ = 10  \\
\hline
23.13  & 23.23  & 23.32  & 22.66 & 21.22 \\
\hline
\end{tabular}
\end{center}
\end{table}

More interestingly, we visualize the patch-wise weight maps of joint SR results in Fig. \ref{fig:temple} - \ref{fig:chip}, using heat maps, as in Fig. \ref{fig:map}. The ($i,j$)-th pixel in the weight map denote the final weight of $\mathbf{X}_{ij}$ when the joint SR reaches a stable solution. All weights are normalized between [0,1], by the form of sigmoid function: $\frac{1}{1+ \omega(\mathbf{\alpha}_{ij}, \mathbf{X}_{ij}^E)}$, for visualization purpose. A larger pixel value in the weight maps denote a smaller weight and thus a higher emphasis on external examples, and vice versa. For \textit{Temple} image, Fig. \ref{fig:map} (a) clearly manifests that self examples dominate the SR of the temple building that is full of texture patterns. Most regions of Fig. \ref{fig:map} (b) are close to 0.5, which means that  $\omega (\mathbf{\alpha}_{ij}, \mathbf{X}_{ij}^E)$ is close to 1 and external and internal examples have similar performances on most patches. However, internal similarity makes more significant contributions in reconstructing the brick regions, while external examples works remarkably better on the irregular contours of forests. Finally, the \textit{Chip} image is an example where external examples have advantages on the majority of patches. Considering self examples prove to create artifacts here (see Fig. \ref{fig:chip} (c) (d)), they are avoided in joint SR by the adaptive weights.




\subsection{SR Beyond Standard Definition: From HD Image to UHD Image}

In almost all SR literature, experiments are conducted with Standard-Definition (SD) images (720 $\times$ 480 or 720 $\times$ 576 pixels) or smaller. The High-Definition (HD) formats:  720p (1280 $\times$ 720 pixels) and 1080p (1920 $\times$ 1080 pixels) have become popular today. Moreover, Ultra High-Definition (UHD) TVs are hitting the consumer markets right now with the 3840 $\times$ 2160 resolution. It is thus quite interesting to explore whether SR algorithms established on SD images can be applied or adjusted for HD or UHD cases. In this section, we upscale HD images of 1280 $\times$ 720 pixels to UHD results of 3840 $\times$ 2160 pixels, using competitor methods and our joint SR algorithm. 


\begin{figure*}[htbp]
\centering
\begin{minipage}{0.90\textwidth}
\centering \subfigure[Full SHD image] {
\includegraphics[width=\textwidth]{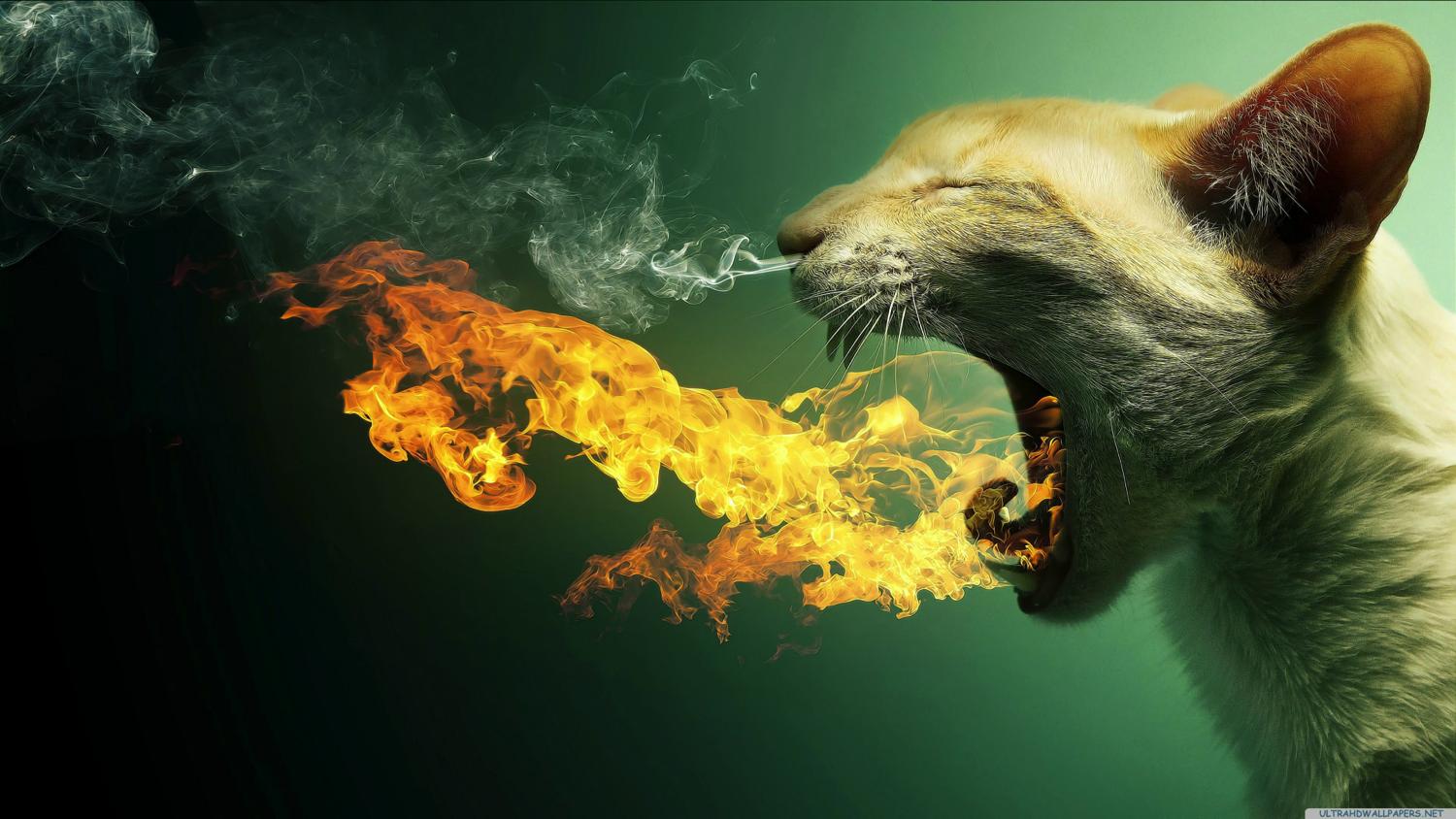}
}\end{minipage}
\begin{minipage}{0.45\textwidth}
\centering \subfigure[Local region from SR result by BCI \newline PSNR = 24.14 dB, SSIM = 0.9701] {
\includegraphics[width=\textwidth]{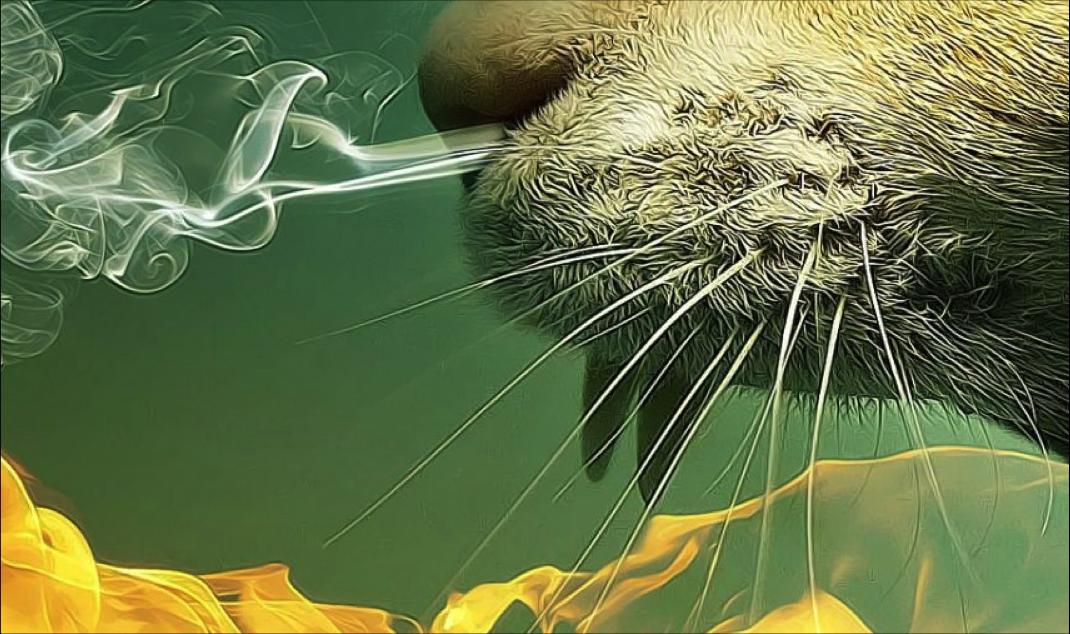}
}
\subfigure [Local region from SR result by CSC \newline PSNR = 25.32 dB, SSIM = 0.9618] {
\includegraphics[width=\textwidth]{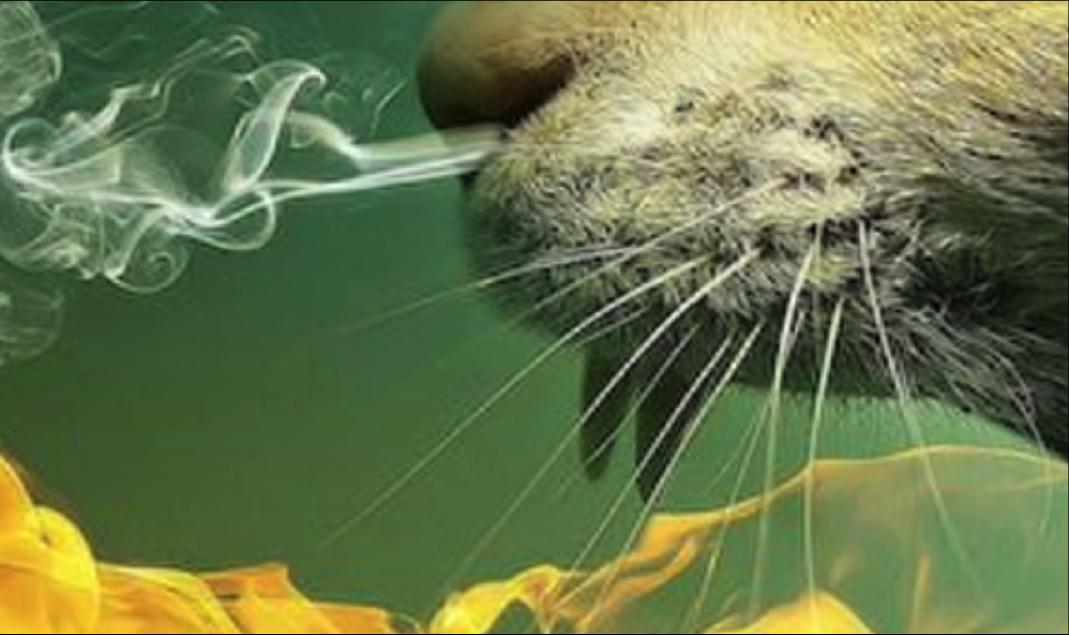}
}\end{minipage}
\begin{minipage}{0.45\textwidth}
\centering \subfigure [Local region from SR result by EPI \newline PSNR = 23.58 dB, SSIM = 0.9656] {
\includegraphics[width=\textwidth]{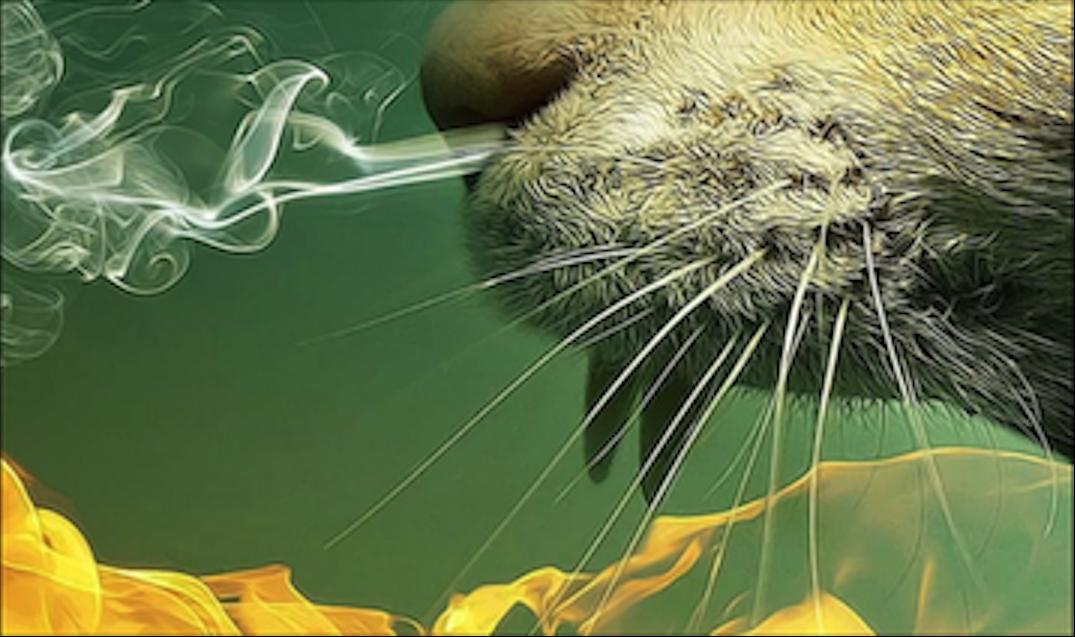}
}
\subfigure [Local region from SR result by JSR \newline PSNR = 25.82 dB, SSIM = 0.9746] {
\includegraphics[width=\textwidth]{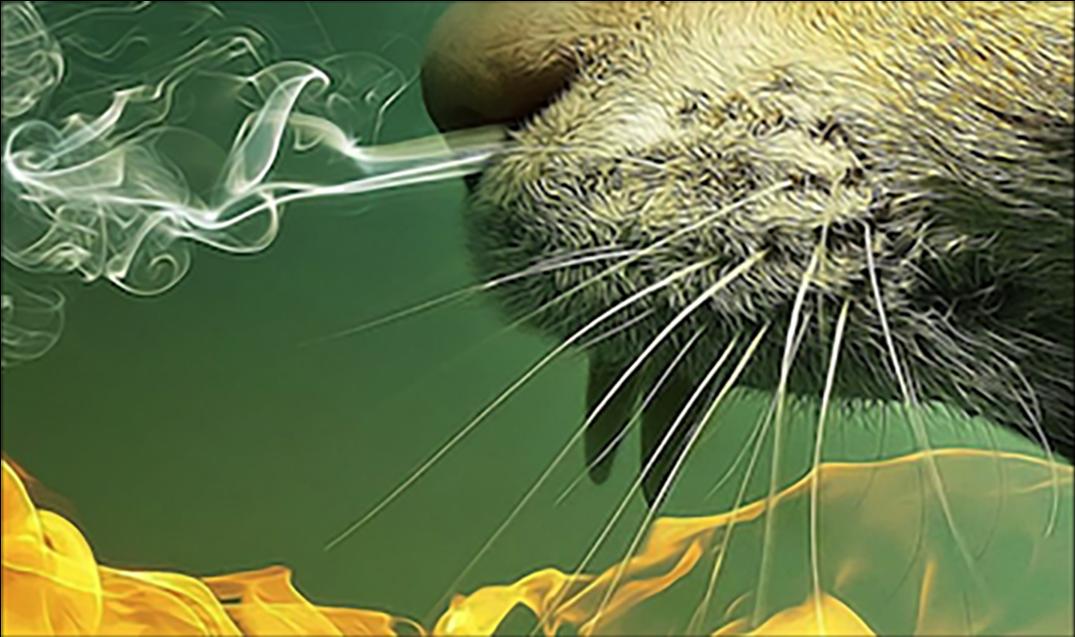}
}\end{minipage}
\caption{3$\times$ SR results of the \textit{Leopard} image (local region displayed).}
\label{fig:Lion}
\end{figure*}



Since most HD and UHD images typically contain much more diverse textures and a richer collection of fine structures than SD images, we enlarge the patch size from $5 \times 5$ to $25 \times 25$ (the dictionary pair is therefore re-trained as well) to capture more variations, meanwhile increasing the overlapping from one pixel to five pixels to ensure enough spatial consistency. Hereby JSR is compared with its two ``component'' algorithms, i.e., CSC and EPI.  We choose several challenging SHD images (3840 $\times$ 2160 pixels) with very cluttered texture regions, downsampling them to HD size (1280 $\times$ 720 pixel) on which we apply the SR algorithm with a factor of 3. In all cases, our results are consistently sharper and clearer. The SR results (zoomed local regions) of the \textit{Leopard} image are displayed in Fig. \ref{fig:Lion} for examples, with the PSNR and SSIM measurements of full-size results.






\subsection{Subjective Evaluation}
\begin{figure}[t]
\center
    \includegraphics[angle=0,width=70mm,clip]{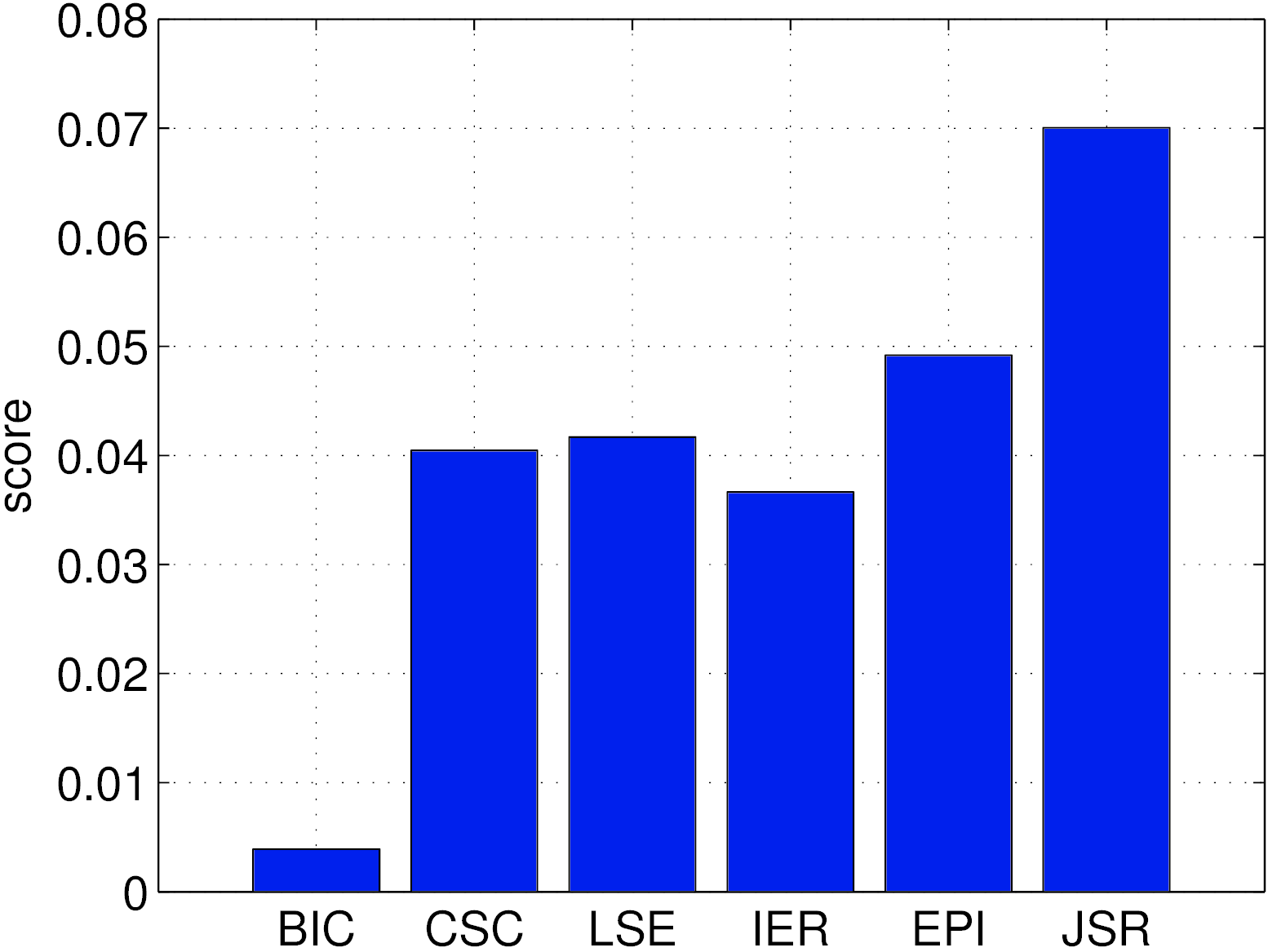}
\caption{Subjective SR quality scores for different methods. The ground truth has score 1.}
\vspace{-3mm}
\label{fig:subeval}
\end{figure}

We conduct an online \textit{subjective evaluation} survey \footnote {\url{http://www.ifp.illinois.edu/~wang308/survey}} on the quality of SR results produced by all different methods in Section 4.2. Ground truth HR images are also included when they are available as references. Each participant of the survey is shown a set of HR image pairs obtained using two different methods for the same LR image. For each pair, the participant needs to decide which one is better than the other in terms of perceptual quality. The image pairs are drawn from all the competitive methods randomly, and the images winning the pairwise comparison will be compared again in the next round, until the best one is selected. We have a total of 101 participants giving 1,047 pairwise comparisons, over six images which are commonly used as benchmark images in SR, with different scaling factors (\textit{Kid}$\times4$, \textit{Chip}$\times4$, \textit{Statue}$\times4$, \textit{Leopard}$\times3$, \textit{Temple}$\times3$ and \textit{Train}$\times3$). We fit a Bradley-Terry \cite{Bradley52rank} model to estimate the subjective scores for each method so that they can be ranked. More experiment details are included in our Appendix. Figure \ref{fig:subeval} shows the estimated scores for the six SR methods in our evaluation. As expected, all SR methods receive much lower scores compared to ground truth (set as score 1), showing the huge challenge of the SR problem itself. Also, the bicubic interpolation is significantly worse than others. The proposed JSR method outperforms all other state-of-the-art methods by a large margin, which proves that JSR can produce more visually favorable HR images by human perception.


\section{Conclusion}
This paper presents a joint single image SR model, by learning from both external and internal examples. We define the two loss functions by sparse coding and epitomic matching, respectively, and construct an adaptive weight to balance the two terms. Experimental results demonstrate that joint SR outperforms existing state-of-the-art methods for various test images of different definitions and scaling factors, and is also significantly more favored by user perception. We will further integrate dictionary learning into the proposed scheme, as well as reducing its complexity.


\appendix
 
\textbf{1. Epitomic Matching Algorithm}
 
 We assume an epitome $\ve$ of size ${M_e} \times {N_e}$, for an input image of size $M \times N$, where ${M_e} < M$ and ${N_e} < N$. Similarly to GMMs, $\ve$ contains three parameters \cite{JojicFK03,Ni09,ChuYLCH10}: $\semean$, the Gaussian mean of size ${M_e} \times {N_e}$; $\sevar$, the Gaussian variance of size ${M_e} \times {N_e}$; and $\vpi$,  the mixture coefficients. Suppose $Q$ densely sampled, overlapped patches from the input image, i.e. $\{\vZ_k\}^Q_{k=1}$. Each $\vZ_k$ contains pixels with image coordinates $\vS_k$, and is associated with a hidden mapping $\vT_k$ from $\vS_k$ to the epitome coordinates. All the $Q$ patches are generated independently from the epitome and the corresponding hidden mappings as below:

\begin{align} \label{eq:patchfromT1}
&\prod_{k=1}^Q p(\{\vZ_k\}^Q_{k=1} | \{\vT_k\}^Q_{k=1}, \ve) = \prod_{k=1}^Q p(\vZ_k | \vT_k, \ve)
\end{align}

The probability $p(\vZ_k | \vT_k, \ve)$ in (\ref{eq:patchfromT1}) is computed by the Gaussian distribution where the Gaussian component is specified by the hidden mapping $\vT_k$. $\vT_k$ behaves similar to the hidden variable in the traditional GMMs.

\begin{figure}[htbp]
\begin{center}
\includegraphics[scale=.3]{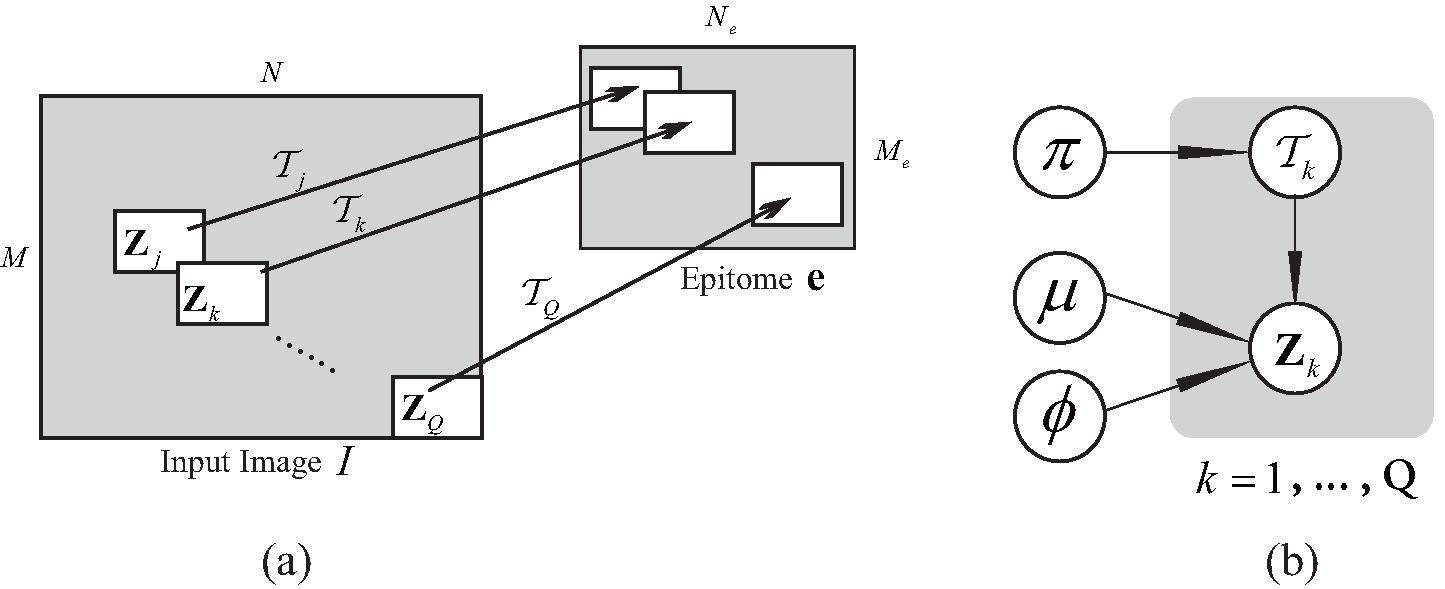}
\end{center}
   \caption{(a) The hidden mapping $\vT_k$ maps the image patch $\vZ_k$ to its corresponding patch of the same size in $\ve$, and $\vZ_k$ can be mapped to any possible epitome patch in accordance with $\vT_k$. (b) The epitome graphical model}
\label{fig:hiddenmapping}
\end{figure}

Figure~\ref{fig:hiddenmapping} illustrates the role that the hidden mapping plays in the epitome as well as the graphical model illustration for epitome. With all the above notations, our goal is to find the epitome $\ve$ that maximizes the log likelihood function ${{\ve}} = \mathop {\arg \max }\limits_{\hat\ve} \log p\left( {\{\vZ_k\}^Q_{k=1}|\hat\ve} \right)$, which can be solved by the Expectation-Maximization (EM) algorithm \cite{JojicFK03,Yang14}. 

The Expectation step in the EM algorithm which computes the posterior of all the hidden mappings accounts for the most time consuming part of the learning process. Since the posterior of the hidden mappings for all the patches are independent of each other, they can be computed in parallel. Therefore, the learning process can be significantly accelerated by parallel computing.

With the epitome $\ve_{\mathbf{Y'}}$ learned from the smoothed input image $\mathbf{Y'}$, the location of the matching patch in the epitome $\ve_{\mathbf{Y'}}$ for each patch $\mathbf X_{ij}^{'E}$  is specified by the most probable hidden mapping for $\mathbf X_{ij}^{'E}$:
\begin{align} \label{eq:mostprobT}
&\vT^*_{ij} = \mathop {\arg \max }\limits_{\vT_{ij}} p\left( {\vT_{ij}|{\mathbf X_{ij}^{'E}},\ve } \right)
\end{align}
The top $K$ patches in $\mathbf{Y'}$ with large posterior probabilities $p\left( {\vT^*_{ij}|{\cdot},\ve} \right)$ are regarded as the candidate matches for the patch $\mathbf{X'}_{ij}$, and the match $\mathbf{Y}_{mn}'$ is the one in these $K$ candidate patches which has minimum Sum of Squared Distance (SSD) to $\mathbf X_{ij}^{'E}$. Note that the indices of the $K$ candidate patches in $\mathbf{Y'}$ for each epitome patch are pre-computed and stored when training the epitome $\ve_{\mathbf{Y'}}$ from the smoothed input image $\mathbf{Y'}$, which makes epitomic matching efficient.


EPI significantly reduces the artifacts and produces more visually pleasing SR results by the dynamic weighting (\ref{eq:weighted-frequency}), compared to the local NN matching method {\cite{Fattal2010}}.

\textbf{2. Subjective Review Experiment}

The methods under comparison include BIC, CSC, LSE, IER, EPI, JSR.
Ground truth HR images are also included when they are available as references.
Each of the human subject participating in the evaluation is shown a set of HR image pairs obtained
using two different methods for the same LR image.
For each pair, the subject needs to decide which one is better than the other in terms of perceptual quality.
The image pairs are drawn from all the competitive methods randomly,
and the images winning the pairwise comparison will be compared again in the next round until the best one is selected.

We have a total of 101 participants giving 1,047 pairwise comparisons over 6 images with different scaling factors
(``Kid''$\times4$, ``Chip''$\times4$, ``Statue''$\times4$, ``Leopard''$\times3$, ``Temple''$\times3$ and ``Train''$\times3$).
Not every participant completed all the comparisons but their partial responses are still useful.
All the evaluation results can be summarized into a $7{\times}7$ winning matrix $\mathbf{W}$ for 7 methods (including ground truth),
based on which we fit a Bradley-Terry \cite{Bradley52rank} model to estimate
the subjective score for each method so that they can be ranked.
In the Bradley-Terry model, the probability that an object $X$ is favored over $Y$ is assumed to be
\begin{equation}
\label{eq:bt}
    p(X \succ Y) = \frac{e^{s_X}}{e^{s_X}+e^{s_Y}} = \frac{1}{1+e^{s_Y-s_X}} ,
\end{equation}
where $s_X$ and $s_Y$ are the subjective scores for $X$ and $Y$.
The scores $\mathbf{s}$ for all the objects can be jointly estimated
by maximizing the log likelihood of the pairwise comparison observations:
\begin{equation}
\label{eq:bt_mle}
    \max_{\mathbf{s}} \sum_{i, j} w_{ij}\log \left(\frac{1}{1+e^{s_j-s_i}}\right) ,
\end{equation}
where $w_{ij}$ is the $(i, j)$-th element in the winning matrix $\mathbf{W}$,
representing the number of times when method $i$ is favored over method $j$.
We use the Newton-Raphson method to solve Eq.~\eqref{eq:bt_mle} and set the score for
ground truth method as 1 to avoid the scale issue.

Fig.~\ref{fig:subeval} shows the estimated scores for six SR methods in our evaluation.
As expected, all the SR methods have much lower scores than ground truth, showing the great challenge in SR problem.
Also, the bicubic interpolation is significantly worse than other SR methods.
The proposed JSR method outperforms other previous state-of-the-art methods by a large margin,
which verifies that JSR can produce visually more pleasant HR images than other approaches.

%

\end{document}